\documentclass{article}

\usepackage{microtype}
\usepackage{graphicx}
\usepackage{subfigure}
\usepackage{booktabs} 
\usepackage{amsmath}
\usepackage{multirow}
\usepackage{pdfpages}
\usepackage{amsfonts}

\usepackage{hyperref}

\usepackage[accepted]{icml2021}

\icmltitlerunning{Evolving Attention with Residual Convolutions}

\begin{document}

\twocolumn[
\icmltitle{Evolving Attention with Residual Convolutions}





\begin{icmlauthorlist}
\icmlauthor{Yujing Wang}{pku,msra}
\icmlauthor{Yaming Yang}{msra}
\icmlauthor{Jiangang Bai}{pku,msra}
\icmlauthor{Mingliang Zhang}{pku,msra} \\

\icmlauthor{Jing Bai}{msra}
\icmlauthor{Jing Yu}{cas}
\icmlauthor{Ce Zhang}{eth}
\icmlauthor{Gao Huang}{thu}
\icmlauthor{Yunhai Tong}{pku}
\end{icmlauthorlist}

\icmlaffiliation{pku}{Peking University}
\icmlaffiliation{msra}{Microsoft Research}
\icmlaffiliation{cas}{Institute of Information Engineering, Chinese Academy of Sciences}
\icmlaffiliation{eth}{ETH Zurich}
\icmlaffiliation{thu}{Tsinghua University}

\icmlcorrespondingauthor{Yujing Wang}{yujwang@pku.edu.cn}
\icmlcorrespondingauthor{Yaming Yang}{yayaming@microsoft.com}


\vskip 0.3in
]



\printAffiliationsAndNotice{}  

\begin{abstract}
Transformer is a ubiquitous model for natural language processing and has attracted wide attentions in computer vision. The attention maps are indispensable for a transformer model to encode the dependencies among input tokens. However, they are learned independently in each layer and sometimes fail to capture precise patterns.
In this paper, we propose a novel and generic mechanism based on evolving attention to improve the performance of transformers. On one hand, the attention maps in different layers share common knowledge, thus the ones in preceding layers can instruct the attention in succeeding layers through residual connections. On the other hand, low-level and high-level attentions vary in the level of abstraction, so we adopt convolutional layers to model the evolutionary process of attention maps.
The proposed evolving attention mechanism achieves significant performance improvement over various state-of-the-art models for multiple tasks, including image classification, natural language understanding and machine translation.
\end{abstract}

\section{Introduction}
Transformer~\citep{vaswani2017attention} is the state-of-the-art architecture for sequential modeling which achieves superior performances in various applications, such as natural language understanding~\citep{devlin2018bert}, image generation~\citep{parmar2018image} and time-series forecasting~\citep{li2019enhancing}. The performance of a transformer model mainly depends on its capability of inducing reasonable attentions between input tokens. However, as illustrated by some previous works~\citep{tang2018self,jain2019attention}, the attention maps captured by vanilla attention layers are not always effective and explainable. To cope with this problem, recent efforts concatenated self-attentions with convolutional layers to obtain better image or text representations~\citep{bello2019attention,wu2020lite}, whereas the attention map itself was not ameliorated.
In this paper, we consider another question, {\em can we improve the learning of attention maps via a dedicated model architecture design?} As we will illustrate, it is possible to improve the quality of attention maps by an evolving mechanism based on a chain of convolutional modules.

\begin{figure}[t]
	\centering
        \includegraphics[width=0.98\linewidth]{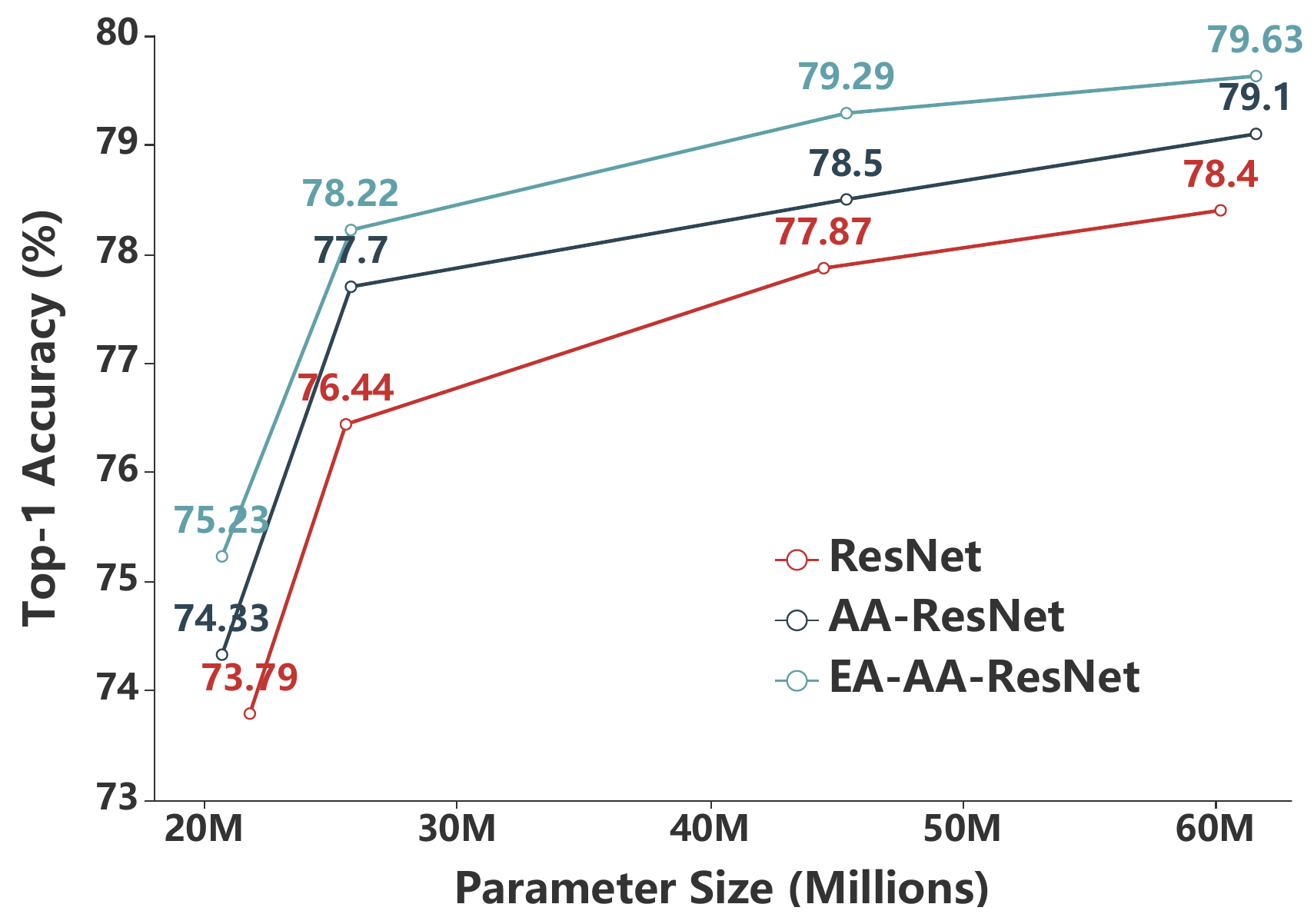}
        \label{fig:imagenet}
         \caption{Comparison of different model architectures for ImageNet classification}
\end{figure}

In a vanilla transformer, the attention maps in each layer are learned independently, which do not have good generalization ability. Intuitively, one can simply share attention maps among layers, but it is not effective as different layers may require attention structures from different abstraction levels. For instance, in image classification, a lower layer usually focuses on the relations within similar colors and textures, while a higher layer needs to reason about dependencies between components. Our motivation is to design a dedicated module to improve the quality of attention maps in an evolutionary process. Therefore, we directly bridge the attention maps from different layers through residual connections. Moreover, we adopt convolutional layers to capture the evolution of attention patterns, as this inductive bias emphasizes local details and produces more precise attention maps by reasoning on previous ones.

To this end, we propose Evolving Attention (EA-) Transformer, which guides the learning of attention maps via a chain of residual convolutional modules coupled with the transformer architecture. In each block, EA-Transformer takes all attention maps generated by the previous block as a multi-channel image. Then, with 2D-convolution over that image, the attention maps for the current block can evolve from previous ones efficiently. As such, the generic patterns of inter-token dependencies are shared across all blocks, and the attention maps are adapted to an appropriate abstraction level for each layer.
\begin{figure*}[t]
	\centering
        \includegraphics[width=0.95\linewidth]{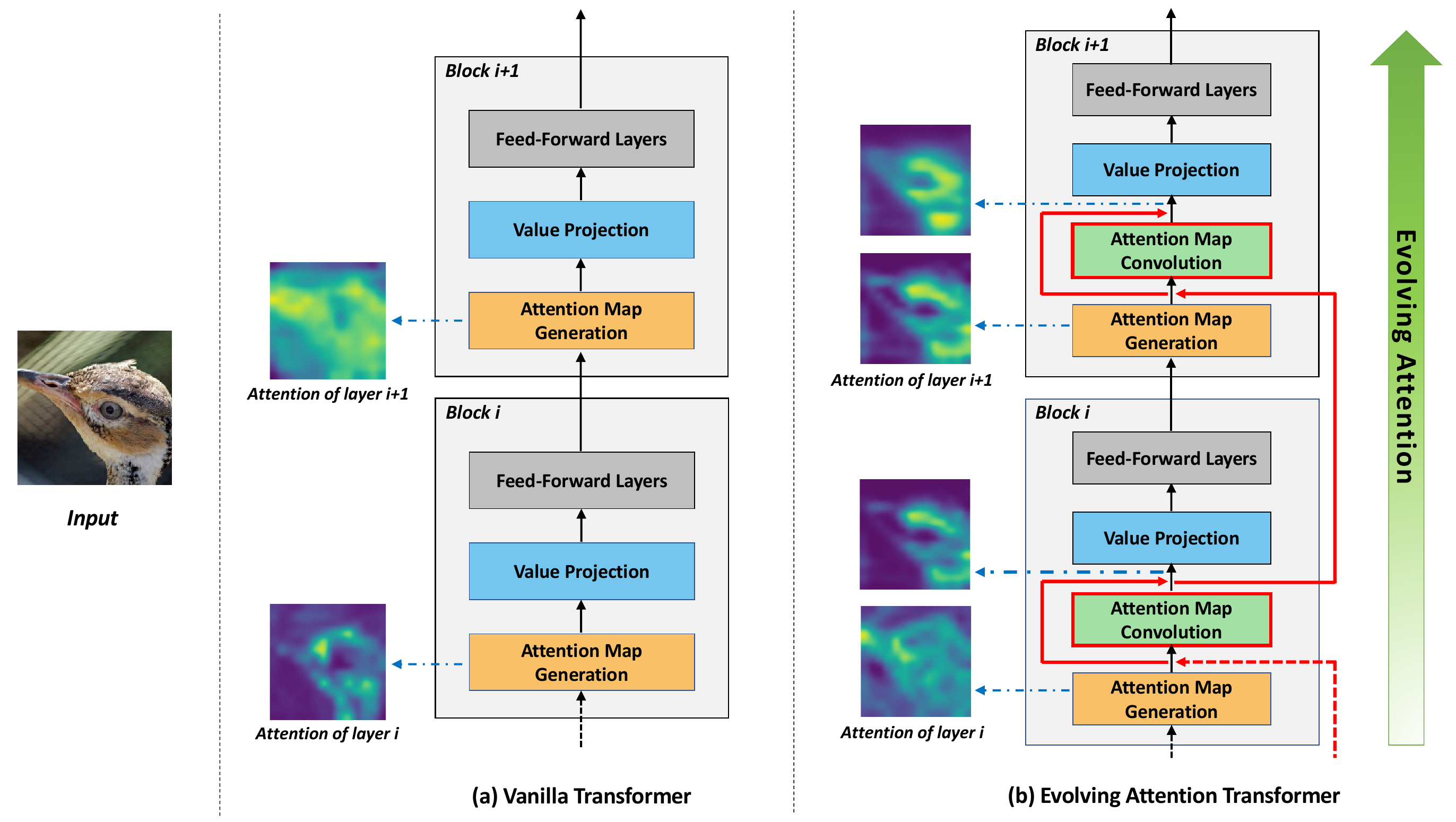}
         \caption{Architecture overview. \small The red lines denote residual connections, and the attention maps from the 17th and 18th blocks of an exemplar case are presented. The vanilla Transformer obtains very broad and vague attention maps at the 18th block. Instead, Evolving Attention Transformer generates reasonable attention maps at both blocks, and there is a clear evolutionary trend across successive layers. }
        \label{fig:overview}
\end{figure*}

As illustrated by a case of ImageNet classification in Figure \ref{fig:overview}, the attention maps learned by EA-Transformer correctly highlight the structure of a bird with the help of convolution-based attention evolution. Especially, the residual links between attention maps facilitate the information flow of inter-token relations. Meanwhile, the convolutional module imitates an evolutionary mechanism and guides the self-attention layer to induce better attention maps. In contrast, the vanilla transformer learns each layer separately and sometimes produces vague attention structures.

We apply the generic idea of evolving attention to multiple state-of-the-art models, including Transformer~\citep{vaswani2017attention}, BERT~\citep{devlin2018bert} and Attention Augmented (AA-) ResNet~\citep{bello2019attention}. Experimental results demonstrate the superiority of evolving attention for various tasks in computer vision and natural language processing domains. As shown in Figure \ref{fig:imagenet}, for ImageNet classification, it consistently improves the accuracy of AA-ResNet with different model capacities. AA-ResNet is a strong SOTA which encapsulates self-attentions and convolutions jointly for image representation. We also examine the generality of evolving attention in BERT-style pre-trained models. Impressively, the average GLUE scores are lifted by 2.4, 1.1, 1.6 and 0.8 points from BERT-Base, T5-Base, BERT-Large, and RoBERTa-Large respectively.


The contributions of this paper are highlighted as follows.
\begin{itemize}
    \item
    We propose a novel evolving attention mechanism augmented by a chain of residual convolutional modules. To the best of our knowledge, this is the first work that considers attention maps as multi-channel images for pattern extraction and evolution, which sheds new lights on the attention mechanism.
    \item
    Extensive experiments have demonstrated consistent improvement in various natural language and computer vision tasks.
    As indicated by extensive analysis, both residual connections and convolutional inductive bias are beneficial to produce better attention maps.
     \item
    The proposed evolving attention mechanism is generally applicable for attention-based architectures and has further impacts on a broader range of applications.
\end{itemize}

\section{Related Work}
Transformer is first introduced by \citet{vaswani2017attention} for machine translation and then widely adopted in numerous tasks in natural language~\citep{devlin2018bert}, computer vision~\citep{parmar2018image, parmar2019stand} and time-series~\citep{li2019enhancing} domains. Transformer is solely composed of self-attention and feed-forward layers. It is much more parallelizable than Recurrent Neural Networks (RNNs) and demonstrates superiority in large-scale training scenarios. Notably, the text representation model, BERT~\citep{devlin2018bert}, is based on an architecture of deep bidirectional Transformer. After pre-trained on a large-scale language corpus, BERT can be fine-tuned with just one additional output layer to create state-of-the-art performance for a wide range of text-related applications.

The assumption behind Transformer is that the intra-sequence relationships can be captured automatically through self-attention. However, in practice, it is questionable if a self-attention layer learns reasonable dependencies among input tokens. Many endeavors are trying to analyze the attention maps generated by the attention mechanism. \citet{raganato2018analysis} analyze the Transformer model for machine translation and show that some attention heads are able to capture certain relations implicitly: lower layers tend to learn more about syntax while higher layers tend to encode more about semantics. \citet{tang2018self} suggest that the ability of inducing syntactic relations for a Transformer model is weaker than its recurrent neural network counterpart.
\citet{tay2020synthesizer} argue that explicit token-token interaction is not important and replace dot-product attention with synthesized attention maps. Moreover, there is a debate on whether or not the intermediate representations offered by attention mechanisms are useful to explain the reasons for a model’s prediction~\citep{jain2019attention,wiegreffe2019attention}. In short, the attention maps induced by existing attention mechanisms are not good enough. Besides, there are successful attempts to combine convolutional and self-attention layers to enrich image and text representations~\citep{bello2019attention, wu2020lite}. However, to the best of our knowledge, our work is one of the first that takes attention maps as multi-channel images and utilizes a dedicated deep neural network for pattern extraction and evolution. We believe this is a promising direction that deserves more investigations in the future.

Another limitation of Transformer lies in its prohibition for modeling long sequences, as both the memory and computation complexities are quadratic to the sequence length. To address this problem, Reformer~\cite{kitaev2020reformer} utilizes two techniques to improve the efficiency of Transformers: (1) revising dot-product attention with locality-sensitive hashing; and (2) replacing residual layers with reversible ones. Moreover, \citet{gehring2017convolutional} leverage an architecture based entirely on convolutional neural networks for sequence to sequence learning, where the number of non-linearities is fixed and independent of the input length. \citet{parmar2019stand} apply stand-alone self-attention layers to image classification by restricting the attention operations within a local region of pixels. Vision Transformer (ViT)~\cite{dosovitskiy2020image} divides an image into a sequence of patches and utilizes an architecture as closely as possible to the text-based Transformer.
In addition, there are other research directions, including relative positional representations~\citep{shaw2018self}, adaptive masks for long-range information~\citep{sukhbaatar2019adaptive}, tree-based transformer~\citep{shiv2019novel}, and AutoML-based evolved transformer~\citep{so2019evolved}. These works are orthogonal to ours and most of them can benefit from our proposed evolving attention mechanism.

\section{Evolving Attention Transformer}

\subsection{Overview}

The representation of a text sequence can be written as $\mathbf{X} \in \mathbf{R}^{N \times C}$,
where $N$ denotes the sequence length and $C$ is the dimension size. For an image representation, the conventional shape is $(H, W, C)$, where $H, W$ and $C$ denote height, width and channel size of the image respectively. In order to apply a standard Transformer to the image representation, we flatten its shape as $\mathbf{X} \in \mathbf{R}^{N \times C}$, where $N=HW$, and each pixel serves as an individual token in the Transformer model.

A standard Transformer block is composed of a self-attention layer and two position-wise feed-forward layers. The attention map is generated by each self-attention layer separately without explicit interactions among each other. However, as we have argued in the introduction, an independent self-attention layer does not have a good generalization ability to capture the underlying dependencies among tokens. Therefore, we adopt a residual convolutional module that generalizes attention maps in the current layer based on the inherited knowledge from previous layers. The proposed mechanism is named as Evolving Attention (EA).

A transformer architecture augmented by evolving attention is illustrated in Figure \ref{fig:overview}(b). Each Evolving Attention (EA-) Transformer block consists of four modules, including \textit{Attention Map Generation}, \textit{Attention Map Convolution}, \textit{Value Projection}, and \textit{Feed-Forward Layers}. 
The residual connections between attention maps (highlighted by the red lines) are by design to facilitate the attention information flow with some regularization effects. Note that we omit layer norms in the figure for brevity. In the rest of this section, the details of each module will be introduced separately.

\subsection{Attention Map Generation}
Given the input representation $\textbf{X}$, the attention maps can be calculated as follows. First, we compute the query and key matrices for each attention head through linear projections, \textit{i.e.}, $\mathbf{Q} = \mathbf{XW}^Q, \mathbf{K} = \mathbf{XW}^K$, where $\mathbf{Q}$ and $\mathbf{K}$ denote query and key matrices respectively, $\mathbf{W}^Q$ and $\mathbf{W}^K$ are linear projection parameters.
Then, the attention map is derived by a scaled dot-product operation:
\begin{equation}
\begin{aligned}
\label{self-attention}
    \mathbf{A} = & \text{Attention}(\mathbf{X}) = softmax(\frac{\mathbf{Q}\mathbf{K}^\top}{\sqrt{d}}). \\
\end{aligned}
\end{equation}
Here $\mathbf{A}$ denotes the attention map and $d$ is the hidden dimension size.
To inject sequential information into the model, we incorporate positional encoding to the input representation. The positional encoding can be either absolute or relative, and we follow the original implementation for each baseline model.
The absolute positional embedding~\citep{vaswani2017attention} is added to token embedding $\mathbf{X}$ directly.
For relative positional representation~\citep{shaw2018self}, the attention formulation can be re-written as:
\begin{equation}
\begin{aligned}
\label{self-attention}
    \mathbf{A} = & \text{Attention}(\mathbf{X}) = softmax(\frac{\mathbf{Q}\mathbf{K}^\top}{\sqrt{d}} + \mathbf{R}), \\
\end{aligned}
\end{equation}
where $\mathbf{R}=\{\mathbf{r}_{ij}\}$ is the matrix of relative positional encoding. For text data, we have $\mathbf{r}_{ij} = \mathbf{q}_i^T\mathbf{e}_{i-j}$, where $\mathbf{e}_{i-j}$ is a trainable embedding vector in terms of the relative indices for two tokens. For image data, we adopt two separate embedding vectors for height and width~\citep{bello2019attention}.
\begin{equation}
\mathbf{r}_{ij} = \mathbf{q}_i^\top \mathbf{e}^H_{h(j)-h(i)} + \mathbf{q}_i^\top \mathbf{e}^W_{w(j)-w(i)},
\end{equation}
where $\mathbf{q}_i$ is the query representation for the $i^{th}$ pixel, $\mathbf{e}^H$ and $\mathbf{e}^W$ represent for trainable embedding vectors of height and width respectively, $h(i)$ and $h(j)$ are the height indices for the $i^{th}$ and $j^{th}$ pixels, and $w(\cdot)$ denotes the index of width.

\subsection{Attention Map Convolution}
In a vanilla Transformer, the attention maps in each layer are calculated independently without explicit interactions between each other. Instead, in EA-Transformer, we build explicit skip connections between adjacent attention maps.
Assume there are $K$ heads in each layer. Then, we have $K$ output attention maps from the \textit{Attention Map Generation} module. They construct a tensor $A \in \mathbb{R}^{N \times N \times K}$ ($N$ is the sequence length), which can be viewed as a $N \times N$ image with $K$ channels. Taking this as input, we adopt one 2D-convolutional layer with $3 \times 3$ kernels to generalize the attention maps. The output channel is also set to be $K$, so the attention maps of all heads can be generated jointly. We apply a ReLU activation after each 2D-convolution layer to provide non-linearity and sparsity.
Finally, the result attention map is combined with input and fed into a softmax activation layer. Mathematically,
\begin{equation}
\begin{aligned}
\label{eq:combine_ratio}
    & \mathbf{A}^{i}_{input} = \alpha \cdot \mathbf{A}^{i-1}_{logit} + (1 - \alpha) \cdot \text{Attention}(\mathbf{X})^{i}_{logit}, \\
    & \mathbf{A}^{i}_{logit} = \beta \cdot \text{CNN}(\mathbf{A}^{i}_{input}) + (1 - \beta) \cdot \mathbf{A}^{i}_{input}, \\
    & \mathbf{A}^{i} = softmax(\mathbf{A}^{i}_{logit}),
\end{aligned}
\end{equation}
where $\mathbf{A}^{i-1}_{logit}$ is the attention logit matrix from the previous block; $\text{Attention}(\mathbf{X})^{i}_{logit}$ is the logit matrix calculated by the current self-attention block, following equation (2) without softmax; $\mathbf{A}^{i}_{input}$ is the combined matrix after residual connection, which serves as input to the convolutional module. $\text{CNN}$ denotes a 2D-convolutional layer with ReLU activation. $\alpha, \beta \in [0, 1]$ are hyper-parameters for linear combination. In our experiments, the values of $\alpha$ and $\beta$ are chosen empirically on the validation set for each task. 

\subsection{Value Projection and Feed-Forward Layers}
Given the attention map $\mathbf{A}$, the rest of a EA-Transformer block includes value projection and position-wise feed-forward layers that are identical to a standard transformer block. The value projection layer can be formulated as:
\begin{equation}
\begin{aligned}
    \mathbf{H}_k = \mathbf{A}_k\mathbf{X}\mathbf{W}_k^V,   ~~~ \mathbf{H} = (\mathbf{H}_{1} \oplus \mathbf{H}_{2} \oplus ... \oplus \mathbf{H}_{K})\mathbf{W}^O, \\
\end{aligned}
\end{equation}
where $\mathbf{A}_k$ is the attention map for the $k^{th}$ head, $\mathbf{W}_k^V$ is the parameter of value projection, and $\mathbf{H}_k$ is the corresponding representation generated by value projection.
Afterwards, the representations of all heads are concatenated (denoted by $\oplus$) and fed into a linear projection layer with parameter $\mathbf{W}^O$.
At last, the block is finished by two position-wise feed-forward layers:
\begin{equation}
    \text{EA-Transformer}(\mathbf{X}) = \text{ReLU}(\mathbf{H}\mathbf{W}_1 + \mathbf{b}_1)\mathbf{W}_2 + \mathbf{b}_2.
\end{equation}
Conventionally, the dimension of $\textbf{W}_1$ is four times of both $\mathbf{W}^O$ and $\mathbf{W}_2$, forming a bottleneck structure.

\subsection{Convolution for Decoders}
\begin{figure}[t]
	\centering
        \includegraphics[width=\linewidth]{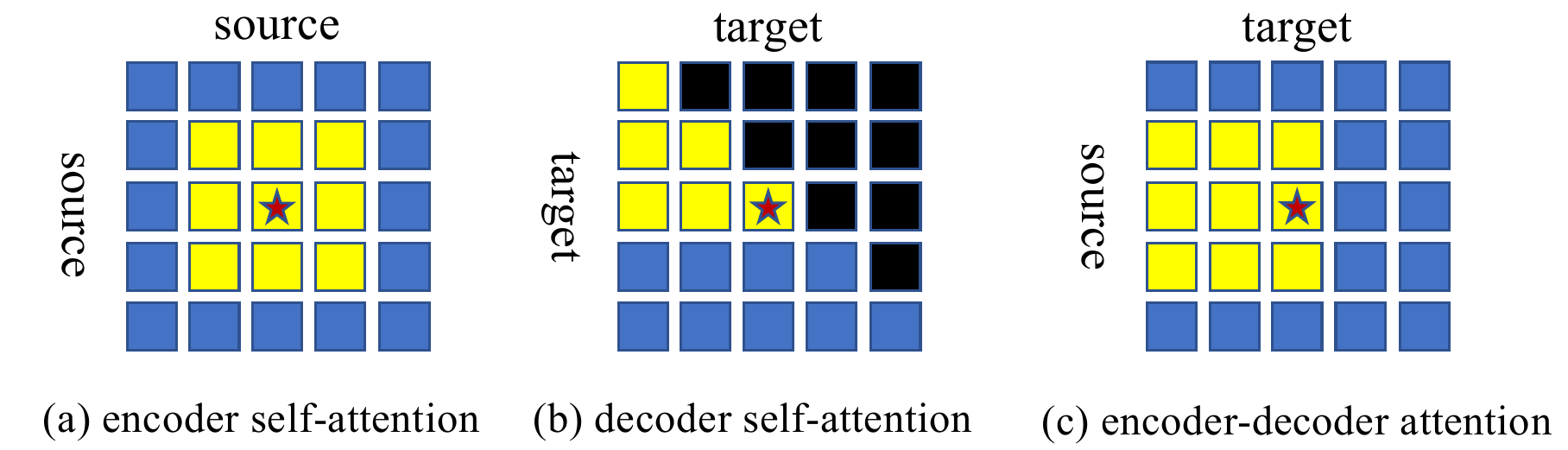}
         \caption{Different convolution strategies for encoder and decoder}
         \label{fig:text}
\end{figure}

In a sequence to sequence transformer network, there are three kinds of attentions, i.e., encoder self-attention, decoder self-attention, and encoder-decoder attention. For the encoder network, we adopt a standard convolution, where the surrounding pixels in a sliding window are taken into consideration. For the decoder part, we need a different convolution strategy to prevent foreseeing subsequent positions. In Figure \ref{fig:text}, we visualize the convolution strategies for three kinds of attention maps, where the current token is identified by a star. The yellow pixels are considered by convolution, while other pixels are not included.
As illustrated by Figure \ref{fig:text}(b), the decoder self-attention only takes upper-left pixels in the convolution. The upper-right pixels (black color) are permanently masked, while other pixels (blue color) are not calculated for the current token. This results in a convolution kernel with a receptive field of 6, which can be implemented as follows: (1) performing standard $3 \times 3$ convolution with masks in the upper-right corner; (2) after convolution, shifting the entire attention matrix by 1 pixel to the bottom and 1 pixel to the right.
As illustrated by Figure \ref{fig:text}(c), the encoder-decoder attention only takes the left pixels in the convolution kernel to prevent information leakage. This can be implemented by a standard $3 \times 3$ convolution with 1 pixel shifting to the right.

\section{Experiments}

\subsection{Image Classification}
\label{exp:ic}

\begin{table}[t]
\centering
\scalebox{0.8}{
    \renewcommand\arraystretch{1.1}
    \centering
    \begin{tabular}{lcccccc}
    \toprule
     \textbf{Model} & \textbf{\#Params} & \textbf{\#FLOPs} & \textbf{Top-1} & \textbf{Top-5}\\
     \midrule
     ResNet-34 & 21.8M & 7.4G & 73.79 & 91.43 \\
     AA-ResNet-34 & 20.7M & 7.1G & 74.33 & 91.92 \\
     \textbf{EA-AA-ResNet-34} & 20.7M& 7.9G & \textbf{75.23} & \textbf{92.35} \\
     \midrule
     ResNet-50 & 25.6M & 8.2G & 76.44 & 93.19 \\
     AA-ResNet-50 & 25.8M & 8.3G & 77.70 & 93.80 \\
     \textbf{EA-AA-ResNet-50} & 25.8M & 8.7G & \textbf{78.22} & \textbf{94.21} \\
     \midrule
     ResNet-101 & 44.5M & 15.6G & 77.87 & 93.89 \\
     AA-ResNet-101 & 45.4M & 16.1G & 78.50 & 94.02 \\
     \textbf{EA-AA-ResNet-101} & 45.4M & 17.2G & \textbf{79.29} & \textbf{94.81} \\
     \midrule
     ResNet-152 & 60.2M&  23.0G & 78.40 & 94.20 \\
     AA-ResNet-152 & 61.6M & 23.8G & 79.10 & 94.60 \\
     \textbf{EA-AA-ResNet-152} & 61.6M & 25.7G & \textbf{79.63} & \textbf{94.85} \\
     \bottomrule
    \end{tabular}
    }
    \caption{Accuracy comparison for ImageNet classification}
    \label{tab:imagenet}
\end{table}

AA-ResNet~\citep{bello2019attention} demonstrated that traditional CNN models could benefit from attention mechanisms in computer vision.
Here we take AA-ResNet as the backbone model for ImageNet classification. AA-ResNet concatenates the image representations computed by self-attentions and convolutional neural networks. 
Our EA-AA-ResNet architecture enhances the attention mechanism by bridging the attention maps from different layers and extracting generic attention patterns through convolutional modules.

\textbf{Settings.}
We follow the experimental protocol of AA-ResNet which adds self-attention to standard ResNet architectures~\citep{he2016deep}. We set $\alpha=0.5$ and $\beta=1.0$ for EA-AA-ResNet-34, while a hyper-parameter analysis and the settings for other architectures can be found in the appendix. All models are trained by 1.28 million training images for 100 epochs on 8 TESLA V100 GPUs. Finally, we report the top-1 and top-5 accuracy on 50k validation images. Major hyper-parameters are as follows: optimizer is SGD with momentum 0.9, batch size is 32 per worker, weight decay is 1e-4. For the first 5 epochs, the learning rate is scaled linearly from 0 to 0.128, and then it is divided by 10 at epoch 30, 60, 80 and 90.

\textbf{Results.} As shown in Table \ref{tab:imagenet}, AA-ResNets consistently outperform corresponding ResNet baselines by a large margin. The proposed EA-AA-ResNets further boost the top-1 accuracy by 1.21\%, 0.67\%, 0.80\% and 0.67\% on top of AA-ResNet-34, -50, -101 and -152 respectively. These numbers are statistically significant under 95\% confidence level. As demonstrated in Figure \ref{fig:imagenet}, the performance enhancement is consistent in different model capacities. Arguably, this is owing to better attention maps induced by the proposed evolving attention mechanism. We will show more evidences in the analysis section.

\begin{table}
\centering
\scalebox{0.8}{
      \begin{tabular}{lcc}
    \toprule
     \textbf{Model} & \textbf{ImageNet Top-1} & \textbf{Top-5}\\
     \midrule
     AA-ResNet-34 & 74.33& 91.92 \\
     EA-AA-ResNet-34 & \textbf{75.23} & 92.35 \\
     $~~$ \textit{w/o Convolution} & 74.34& 91.98  \\
     $~~$ \textit{w/o Skip Connection} & 74.29& 91.85 \\
     $~~$ \textit{with $1 \times 1$ Convolution} & 74.99 & 92.20 \\
     $~~$ \textit{with $5 \times 5$ Convolution} & 75.12 & \textbf{92.55} \\
     \bottomrule
    \end{tabular}
    }
    \caption{Ablation study for ImageNet classification}
    \label{tab:ic_ablation}
\end{table}

\textbf{Ablation Study.}
To understand the importance of each component, we conduct ablation experiments for the EA-AA-ResNet-34 architecture. In Table \ref{tab:ic_ablation}, \textit{w/o Convolution} means removing the attention convolution module from EA-AA-ResNet ($\beta=0$); \textit{w/o Skip Connection} means removing skip connections between adjacent attention maps ($\alpha = 0$); Besides, we also replace standard $3 \times 3$ convolutions with $1 \times 1$ or $5 \times 5$ kernels. As indicated by the results, both convolutions and skip connections are crucial for the final performance. One can also notice a significant accuracy drop using $1 \times 1$ convolutions (analogies to FFN), which indicates that a convolutional inductive bias is beneficial for capturing evolving attention patterns. Meanwhile, $5 \times 5$ convolutions also work fairly well, but we prefer $3 \times 3$ convolutions as they require fewer parameters.

\subsection{Natural Language Understanding}
\label{exp:bert}

\begin{table*}[t]
\centering
    \scalebox{0.75}{
    \renewcommand\arraystretch{1.1}
    \centering
    \begin{tabular}{lccccccccccc}
    \toprule
     \textbf{Model} & \textbf{\#Params} & \textbf{\#FLOPs} & \textbf{Avg} & \textbf{CoLA} & \textbf{SST-2} & \textbf{MRPC} & \textbf{STS-B} & \textbf{QQP} & \textbf{MNLI-m/-mm} & \textbf{QNLI} & \textbf{RTE} \\
     \midrule
     BERT-Base & 109.5M & 6.3G & 80.9 & 51.7 & 93.5 & 87.2/82.1 & 86.7/85.4 & 91.1/89.0 & 84.3/83.7 & 90.4 & 67.2 \\
     \textbf{EA-BERT-Base} & 110.4M & 6.8G & \textbf{83.3} & \textbf{60.0} & \textbf{93.8} & \textbf{89.1/84.2} & \textbf{89.6/89.5} & \textbf{91.6/88.4} & \textbf{85.2/85.5} & \textbf{92.1} & \textbf{69.0} \\
     \midrule
     T5-Base & 220.2M & 9.1G & 83.4 & 53.1 & 92.2 & 92.0/88.7 & 89.1/88.9 & 88.2/91.2 & 84.7/85.0 & 91.7 & 76.9 \\
     Synthesizer & 272.0M & 11.3G & 84.0 & 53.3 & 92.2 & 91.2/87.7 & 89.3/88.9 & 88.6/91.4 & 85.0/84.6 & \textbf{92.3} & 81.2 \\
     \textbf{EA-T5-Base} & 221.2M & 9.9G & \textbf{84.5} & \textbf{53.7} & \textbf{93.1} & \textbf{92.3/89.0} & \textbf{89.6/89.1} & \textbf{88.8/91.9} & \textbf{85.1/85.0} & \textbf{92.3} & \textbf{81.5} \\
     \midrule
     BERT-Large & 335.0M & 12.2G & 83.7 & 60.5 & 94.9 & 89.3/85.4 & 87.6/86.5 & 92.1/89.3 & 86.8/85.9 & 92.7 & 70.1 \\
     RealFormer & 335.0M & 12.2G & 84.3 & 59.8 & 94.0 & \textbf{90.9}/87.0 & \textbf{90.1/89.9} & 91.3/88.3 & 86.3/86.3 & 91.9 & \textbf{73.7}\\
     \textbf{EA-BERT-Large} & 336.7M & 12.9G & \textbf{85.3} & \textbf{62.9} & \textbf{95.2} & \textbf{90.9}/\textbf{89.4} & 89.7/88.2 & \textbf{92.4/90.1} & \textbf{87.9/86.8} & \textbf{93.9} & 72.4 \\
     \midrule
     RoBERTa-Large & 355.0M & 12.7G & 86.4 & 63.8 & 96.3 & 91.0/89.4 & 72.9/90.2 & 92.7/90.1 & 89.5/89.7 & 94.2 & 84.2 \\
     \textbf{EA-RoBERTa-Large} & 356.7M & 13.3G & \textbf{87.2} & \textbf{65.8} & \textbf{96.4} & \textbf{91.8/90.8} & \textbf{73.8/90.3} & \textbf{93.3/90.2} & \textbf{90.5/89.8} & \textbf{95.2} & \textbf{85.0} \\
     \bottomrule
    \end{tabular}
    }
    \caption{Comparison of different models on GLUE benchmark.}
    \label{tab:glue}
\end{table*}

Pre-trained language models like BERT become popular in recent years.
These models are based on bi-directional transformer architectures and pre-trained by a large corpus.
Our solution can be easily plugged into an existing checkpoint of vanilla BERT and achieve significant improvement through continuous training. To demonstrate this advantage, we choose GLUE benchmark~\citep{wang2018glue} for an empirical study.

\textbf{Settings.}
The encoder network of BERT consists of multiple transformer blocks. In EA-BERT, we replace each transformer block with EA-Transformer illustrated in Figure \ref{fig:overview}(b). We load the pre-trained checkpoints of BERT-Base, T5-Base, BERT-Large and RoBERTa-Large directly, and fine-tune them for each downstream task individually on the task-specific training data. The additional parameters introduced by EA-BERT are initialized randomly and trained jointly with other parameters during fine-tuning.
We use the Adam optimizer~\citep{kingma2014adam} with epsilon 1e-8.
The dropout ratio is set as 0.1 empirically. Hyper-parameters are tuned in the following search space on the validation set: learning rate \{1e-4, 1e-5, 2e-5\}, batch size \{8, 16\}, training epochs \{2, 3, 5\}, $\alpha=$ \{0.1, 0.2, 0.4\} and $\beta=$ \{0.1, 0.2, 0.4\}. The values of hyper-parameters chosen for each task will be reported in the appendix.



\textbf{Results.}
The comparison between BERT-style models are shown in Table \ref{tab:glue}.
The T5-Base and BERT-Large models are evaluated on the development set in order to be comparable with existing baselines. Other models are evaluated on the test set.
EA-BERT generally performs better than vanilla BERT in different downstream tasks. Specifically, EA-BERT-Base, EA-T5-Base, EA-BERT-Large and EA-RoBERTa-Large achieve average scores of 83.3, 84.5, 85.0 and 87.2 on the GLUE benchmark, increasing 2.4, 1.1, 1.6 and 0.8 absolute points from the corresponding baselines respectively. The improvement can be achieved by loading an existing checkpoint and fine-tuning the extra parameters with limited training time, which is an appealing advantage in practice.
In addition, it is worth mentioning that EA-BERT-Base boosts the score by 8.1 on CoLA dataset, indicating its superior generalization ability for small datasets.
Moreover, the EA-based models demonstrate more benefits over two concurrent works, Realformer~\cite{he2020realformer} and Synthesizer~\citep{tay2020synthesizer}. Realformer utilizes residual connections over attention maps, but does not exploit convolution-based modules for evolving attention patterns. Synthesizer generates a separate attention map and mixes it with vanilla ones.

\begin{table}
\centering
\scalebox{0.8}{
    \renewcommand\arraystretch{1.1}
    \begin{tabular}{lcccc}
    \toprule
     \textbf{Model} & \textbf{CoLA} & \textbf{SST-2} & \textbf{MRPC} & \textbf{MNLI} \\
     \midrule
     BERT-Base & 51.7 & 93.5 & 87.2/82.1 & 84.3/83.7\\
     EA-BERT-Base & \textbf{60.0} & \textbf{93.8} & \textbf{89.1/84.2} & \textbf{85.2/85.5}\\
     $~$ \textit{w/o Convolution} & 52.1 & 93.6 & 87.6/82.8 & 84.5/84.0\\
     $~$ \textit{w/o Skip Connection} & 52.9 & 93.5 & 87.4/82.4 & 84.3/83.8\\
     $~$ \textit{with $1 \times 1$ Convolution} & 57.5 & 93.6 & 88.4/83.4 & 84.7/84.8\\
     $~$ \textit{with $5 \times 5$ Convolution} & 59.2 & 93.6 & 89.0/83.9 & 85.2/85.4\\
     \bottomrule
    \end{tabular}
}
    \caption{Ablation study for text understanding}
    \label{tab:tu_ablation}
\end{table}

\textbf{Ablation Study}
We perform an ablation study for EA-BERT-Base on four text datasets with different data scales. According to the results in Table \ref{tab:tu_ablation}, the privilege of EA-BERT comes from both convolution-based pattern extraction and residual connections simultaneously. Removing the convolutional module or replacing it with $1 \times 1$ kernel causes a significant performance drop. Skip connections are also advantageous, without which we only have small gains over the vanilla model. Similar to image classification, $5 \times 5$ kernel also generates competitive results. 

\begin{table*}[t]
\centering
\scalebox{0.8}{
    \renewcommand\arraystretch{1.1}
    \centering
    \begin{tabular}{lccccc}
    \toprule
     \textbf{Model} & \textbf{\#Params} & \textbf{\#FLOPs (En-De)} & \textbf{IWSLT’14 De-En} & \textbf{WMT'14 En-De} & \textbf{WMT'14 En-Fr} \\
     \midrule
     Transformer-Lite & 2.48M & 158.31M & 33.32 & 21.11 & 33.22 \\
     \textbf{EA-Transformer-Lite} & 2.49M & 163.46M & \textbf{33.80} & \textbf{21.63} & \textbf{34.12} \\
     \midrule
     Transformer-Base & 44.14M & 2.68G & 34.55 & 27.47  & 40.79 \\
     \textbf{EA-Transformer-Base} & 44.15M & 2.70G & \textbf{35.30}  & \textbf{27.56} & \textbf{41.54} \\
     \bottomrule
    \end{tabular}
    }
    \caption{BLUE scores on machine translation datasets}
    \label{tab:mt}
\end{table*}

\begin{table}[t]
\centering
\scalebox{0.8}{
    \renewcommand\arraystretch{1.1}
    \begin{tabular}{lcccc}
    \toprule
     \textbf{Model} & \textbf{De-En} & \textbf{En-De} & \textbf{En-Fr} \\
     \midrule
     Transformer-Lite & 33.21 & 21.11 & 33.22 \\
     EA-Transformer-Lite & \textbf{33.80} & \textbf{21.63} & \textbf{34.12} \\
     $~~$ \textit{w/o Encoder Convolution} & 32.84 & 21.07 & 33.51 \\
     $~~$ \textit{w/o Decoder Convolution} & 33.56 & 21.45 & 33.72 \\
     $~~$ \textit{w/o Encoder-Decoder Convolution} & 33.59 & 21.41 & 33.73 \\
     $~~$ \textit{w/o Skip Connection} & 33.70 &  21.43 & 34.06 \\
     $~~$ \textit{with $1 \times 1$ Convolution} & 33.15 & 21.20 & 33.52 \\
     $~~$ \textit{with $5 \times 5$ Convolution} & 33.54 & 21.40 & 34.08 \\
     \bottomrule
    \end{tabular}
}
 \caption{Ablation study for machine translation}
 \label{mt_ablation}
\end{table}

\subsection{Machine Translation}
\label{exp:mt}
Machine translation is a common benchmark for testing sequence to sequence architectures. In our experiments, we take three machine translation datasets: IWSLT’14 German-English (De-En), WMT'14 English to German (En-De) and WMT'14 English to French (En-Fr).

\textbf{Settings.}
In an EA-Transformer, the convolution modules are applied to encoder self-attention, decoder self-attention and encoder-decoder attention separately. Skip connections are only used in the encoder network as we find they harm the performance of a decoder. We set $\alpha=0.1, \beta=0.1$ for EA-Transformer-Lite and $\alpha=0.5, \beta=0.1$ for EA-Transformer-Base.
We adopt Adam optimizer with $\beta_1 = 0.9$, $\beta_2 = 0.98$ and an inverse square root learning rate scheduling with linear warm-up. Warm-up step is set to be 4000 and label smoothing is 0.1. For each task, we tune the learning rate from \{1e-4, 5e-4, 1e-3\} and dropout ratio from \{0.1, 0.2, 0.3\}, which will be reported in the appendix.

\textbf{Results.}
We compare Transformer and EA-Transformer with different model capacities. Transformer-Lite~\cite{wu2020lite} is a light architecture where all dimensions are set as 160 to replace a bottleneck structure. Transformer-Base follows the configuration in \citet{vaswani2017attention}, which has six layers for the encoder and 6 layers for the decoder network. It has 8 heads, 512 dimension for normal layers and 2048 dimension for the first FFN layer, forming a bottleneck structure.
As illustratged in Table \ref{tab:mt}, EA-based models achieve consistent improvement for multiple datasets and network architectures while requiring only a few extra parameters and computations.


\textbf{Ablation Study.}
The ablation results of EA-Transformer-Lite are listed in Table \ref{mt_ablation}. First, we remove the convolutional modules for encoder, decoder, and encoder-decoder attentions respectively. As a result, the convolutional module is extremely important for the encoder network. It also has positive effects on decoder self-attention and encoder-decoder attentions. Consistent with previous conclusions, replacing $3 \times 3$ convolutions with $1 \times 1$ kernels leads to a huge decay in BLUE score. Therefore, the locality inductive bias is indefensible for a transformer model to evolve for better attention structures.


\section{Analysis}
\label{analysis}

 \begin{figure*}[t]
  \centering
        \subfigure[Input]{
        \begin{minipage}[b]{0.12\linewidth}
        \includegraphics[width=\linewidth]{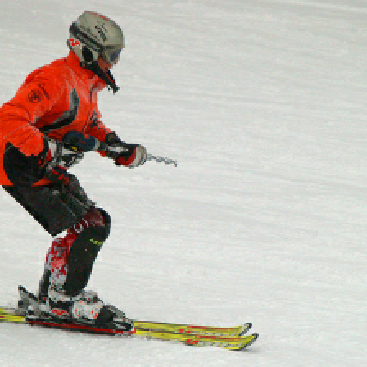}
        \includegraphics[width=\linewidth]{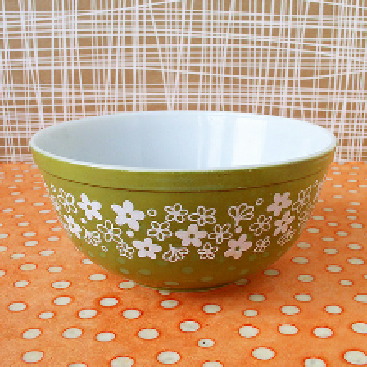}
        \includegraphics[width=\linewidth]{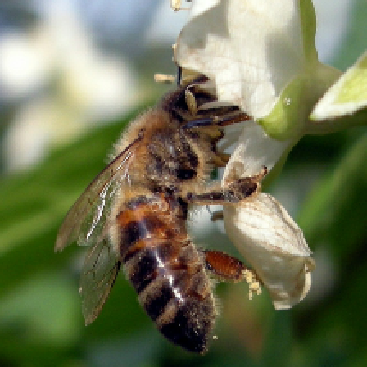}
        \end{minipage}
        }
        \subfigure[AA-16]{
        \begin{minipage}[b]{0.12\linewidth}
        \includegraphics[width=\linewidth]{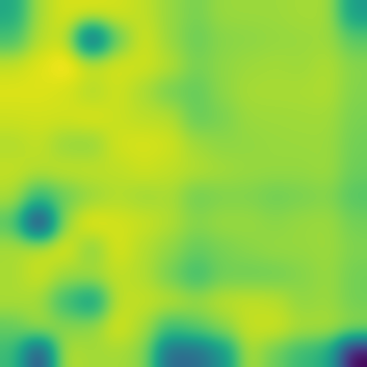}
        \includegraphics[width=\linewidth]{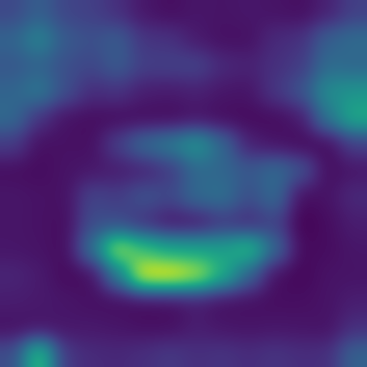}
        \includegraphics[width=\linewidth]{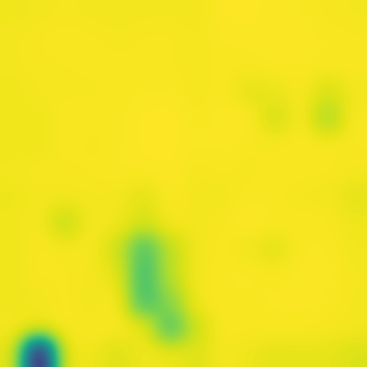}
         \end{minipage}
        }
         \subfigure[AA-17]{
        \begin{minipage}[b]{0.12\linewidth}
        \includegraphics[width=\linewidth]{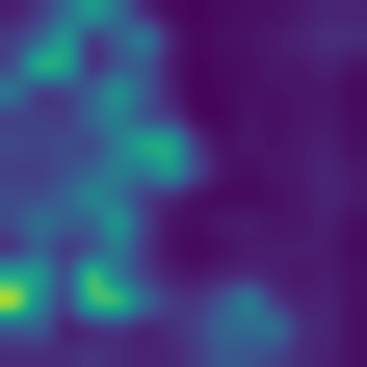}
        \includegraphics[width=\linewidth]{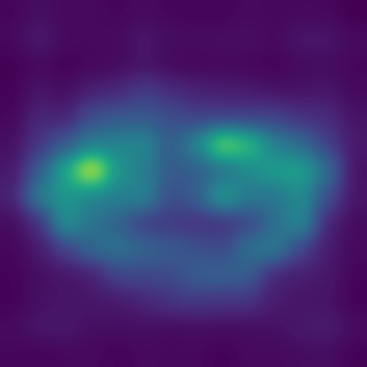}
        \includegraphics[width=\linewidth]{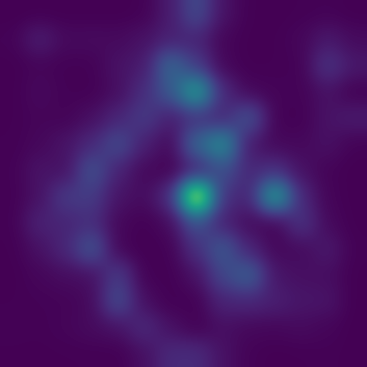}
\end{minipage}}
         \subfigure[AA-18]{
        \begin{minipage}[b]{0.12\linewidth}
        \includegraphics[width=\linewidth]{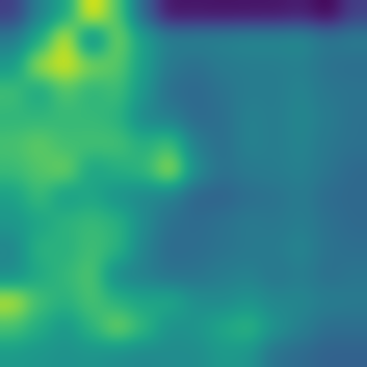}
        \includegraphics[width=\linewidth]{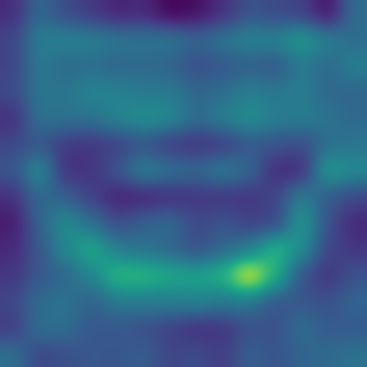}
        \includegraphics[width=\linewidth]{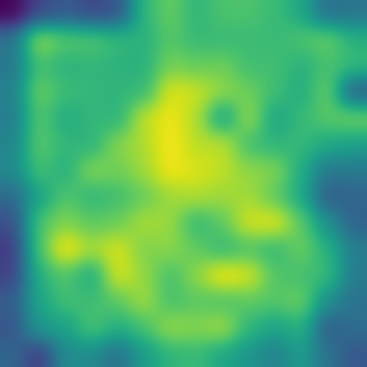}
\end{minipage}}
        \subfigure[EA-AA-16]{
        \begin{minipage}[b]{0.12\linewidth}
        \includegraphics[width=\linewidth]{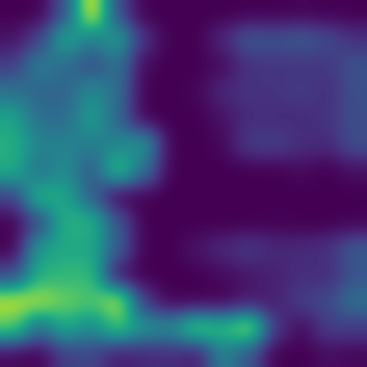}
        \includegraphics[width=\linewidth]{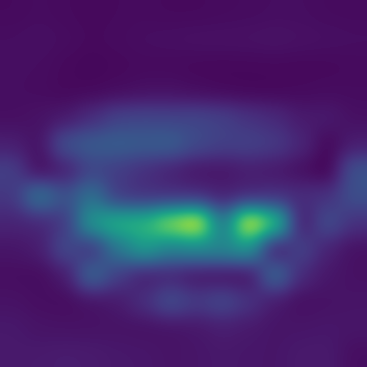}
        \includegraphics[width=\linewidth]{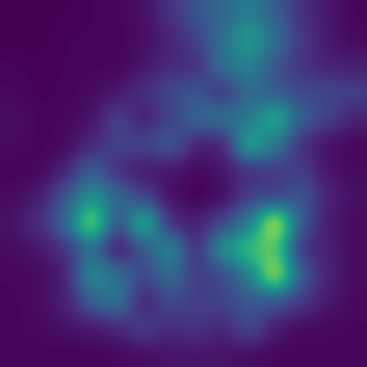}
         \end{minipage}
        }
        \subfigure[EA-AA-17]{
        \begin{minipage}[b]{0.12\linewidth}
        \includegraphics[width=\linewidth]{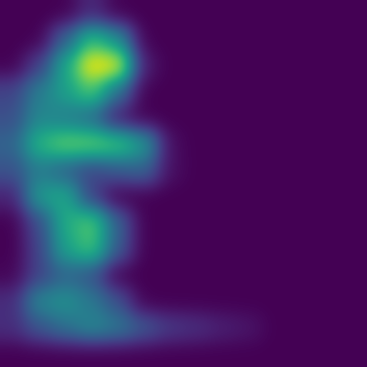}
        \includegraphics[width=\linewidth]{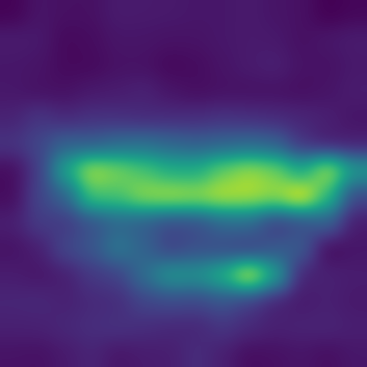}
        \includegraphics[width=\linewidth]{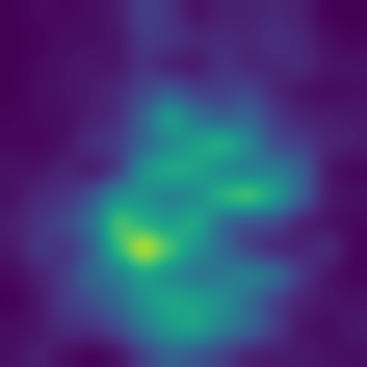}
         \end{minipage}
        }
         \subfigure[EA-AA-18]{
        \begin{minipage}[b]{0.12\linewidth}
        \includegraphics[width=\linewidth]{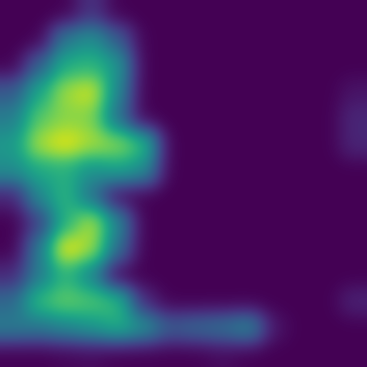}
        \includegraphics[width=\linewidth]{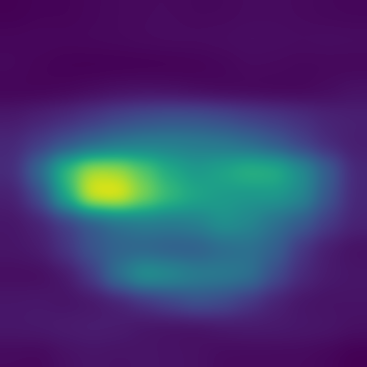}
         \includegraphics[width=\linewidth]{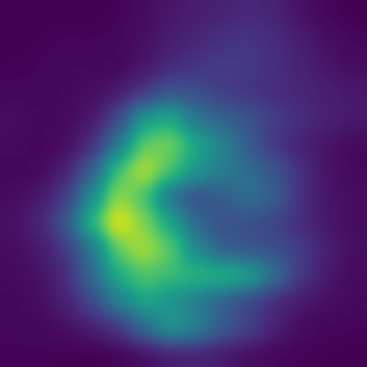}
         \end{minipage}
        }
        \caption{Visualization of exemplar attention maps from the 16th, 17th and 18th layers of AA-ResNet-34 and EA-AA-ResNet-34 models.}
        \label{fig:attention_map}
 \end{figure*}


\subsection{Quality of Attention Maps}

\begin{table}[t]
\centering
\scalebox{0.8}{
    \renewcommand\arraystretch{1.1}
    \begin{tabular}{lcccc}
    \toprule
     \textbf{Models} & \textbf{16th Layer} & \textbf{17th Layer} & \textbf{18th Layer} \\
     \midrule
     AA-ResNet-34 & 24.50 & 31.18 & 17.55 \\
     EA-AA-ResNet-34 & \textbf{31.02} & \textbf{31.71} & \textbf{31.93} \\
     \bottomrule
    \end{tabular}
}
 \caption{Accuracy for attention map classification}
 \label{image_analysis}
\end{table}

To examine the quality of image attention maps, we select the middle attention layers (16th, 17th and 18th) in the 34-layers network for analysis.
We take the attention maps centered on the middle pixel, the shape of which is $N \times N \times K$, where $N=14$ is the image length (after pooling) and $N=8$ is the number of heads. We take these attention maps as direct inputs to another CNN model for classification. If the key structures are better captured in the attention maps, we should get higher accuracy scores for this task. As shown in Table \ref{image_analysis}, AA-ResNet-34 does not generate precise attention maps for the 16th and 18th layers, as the classification accuracy is relatively low. Instead, EA-AA-ResNet-34 induces good attention maps for all three layers, indicating the superiority of evolving attention.

Figure \ref{fig:attention_map} shows three exemplar cases for ImageNet classification, where the attention maps from representative heads of the 16th, 17th and 18th layers are visualized. These layers are at the middle of the network, which correspond to an appropriate abstraction level for visualization. Notably, AA-ResNet prefers to extract broad and vague attention patterns. In contrast, EA-AA-ResNet generates much sharper attention maps, and there exists a clear evolutionary trend in three consecutive layers. For the skier case, the attention map has successfully captured the main object in the 16th layer. Then, the outline becomes much clearer at the 17th layer with the assistance of evolving attention. Finally, the 18th layer is further improved as it identifies a complete skateboard. Other cases demonstrate a similar phenomenon.


\begin{table}[t]
\centering
\scalebox{0.75}{
    \renewcommand\arraystretch{1.1}
    \begin{tabular}{llcccc}
    \toprule
     \textbf{Dataset} & \textbf{Model} & \textbf{Perf.}$\mathbf{\uparrow}$ &\textbf{AUPRC}$\mathbf{\uparrow}$ & \textbf{Comp.}$\mathbf{\uparrow}$ & \textbf{Suff.}$\mathbf{\downarrow}$ \\
     \midrule
     \multirow{2}{*}{\textbf{Movie Reviews}} &
     BERT & 0.970 & 0.280 & 0.187 & 0.093\\
     & EA-BERT & \textbf{0.975} & \textbf{0.313} & \textbf{0.194} & \textbf{0.089}\\
     \midrule
     \multirow{2}{*}{\textbf{FEVER}} &
     BERT & 0.870 & 0.291 & 0.212 & \textbf{0.014}\\
     & EA-BERT &  \textbf{0.886} & \textbf{0.307} & \textbf{0.236} & \textbf{0.014}\\
     \midrule
     \multirow{2}{*}{\textbf{MultiRC}} &
     BERT & 0.655 & 0.208 & 0.213 & -0.079\\
     & EA-BERT & \textbf{0.674} & \textbf{0.221} & \textbf{0.241} & \textbf{-0.089}\\
     \midrule
      \multirow{2}{*}{\textbf{CoS-E}} &
     BERT & 0.487 & 0.544 & 0.223 & 0.143\\
     & EA-BERT & \textbf{0.491} & \textbf{0.552} & \textbf{0.231} & \textbf{0.140} \\
     \midrule
      \multirow{2}{*}{\textbf{e-SNLI}} &
     BERT & 0.960 & 0.399 & 0.177 & 0.396\\
     & EA-BERT & \textbf{0.969} & \textbf{0.534} & \textbf{0.445} & \textbf{0.368}\\
    \bottomrule
    \end{tabular}
}
 \caption{Comparison of text representation models on ERASER benchmark. ``Perf." is accuracy (CoS-E) or F1 (others), AUPRC means Area Under the Precision Recall Curve; ``Comp." and ``Suff." denote comprehensiveness and sufficiency metrics respectively.}
 \label{rationale}
\end{table}

\subsection{Interpretability}
A good model should generate not only correct predictions, but also give faithful reasons to explain the decisions. The ERASER benchmark~\cite{deyoung-etal-2020-eraser} is proposed to evaluate the faithfulness of rationales for text representation models.
Following the original setting, we adopt BERT+LSTM and EA-BERT+LSTM as the text representation models respectively, and utilize a state-of-the-art method, LIME~\citep{ribeiro2016should} to generate rationales for each model.
The experimental results are listed in Table \ref{rationale}. It should be noted that higher comprehensiveness scores and lower sufficiency scores are desired. According to the results, the rationales given by evolving attention-based models are more accurate than those generated by vanilla ones. Thus, the evolving attention mechanism not only boosts the performances of text understanding models, but also improves their interpretability through inducing better attention maps.


\subsection{Learning Curve Comparison}
\begin{figure}[t]
	\centering
        \includegraphics[width=0.95\linewidth]{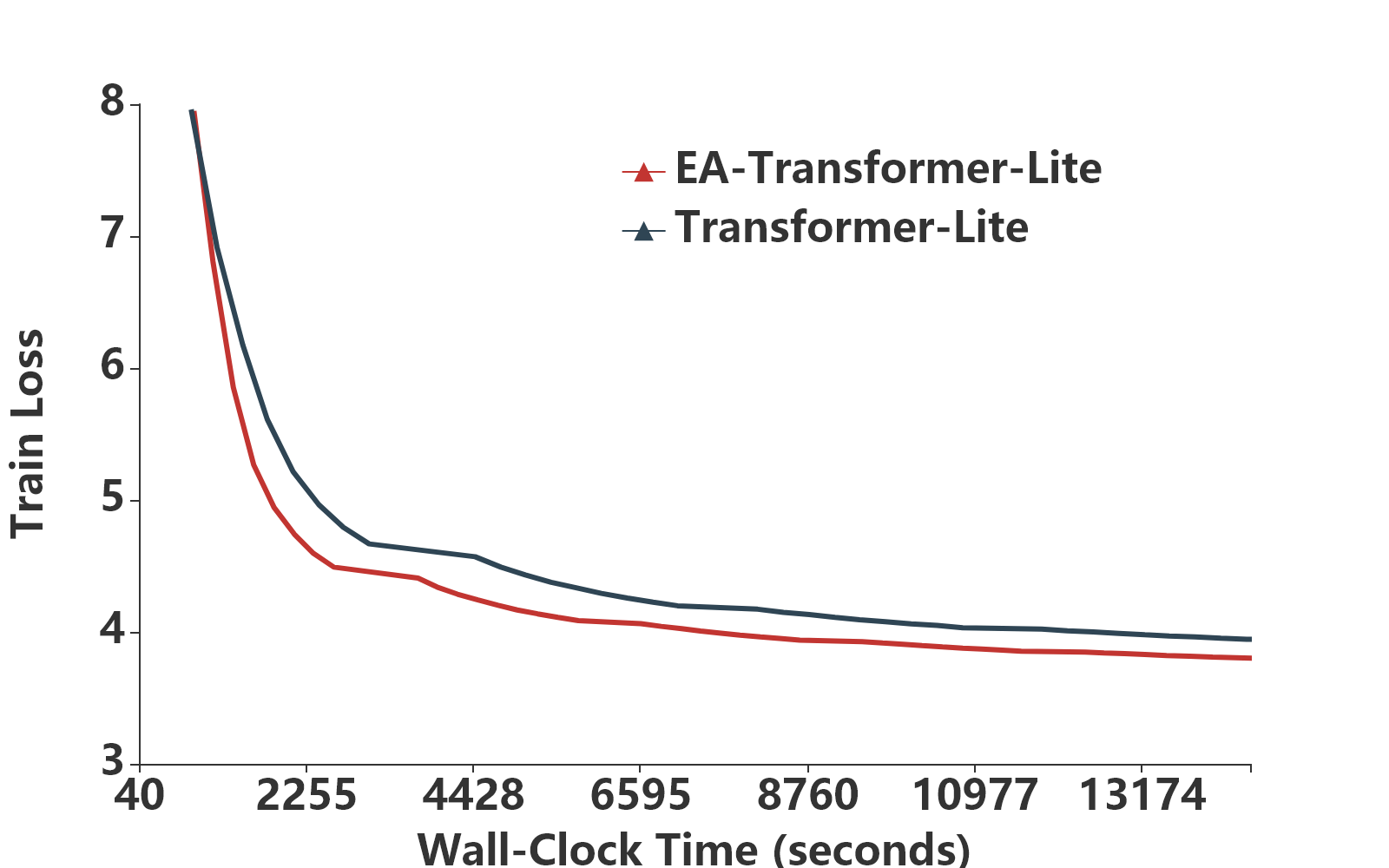}
         \caption{Learning curve comparison on IWSLT’14 De-En machine translation dataset}
         \label{fig:learning_curve}
\end{figure}

To demonstrate the efficiency of EA-Transformer, we further compare the learning curves of Transformer-Lite and EA-Transformer-Lite on IWSLT’14 De-En dataset. As shown in Figure \ref{fig:learning_curve}, EA-Transformer-Lite always achieves lower training loss at the same wall-clock time, although it contains relatively 3\% more FLOPs in each iteration. Finally, EA-Transformer-Lite achieves better training loss and BLUE score at convergence.

\section{Conclusion}
In this paper, we propose a novel mechanism, Evolving Attention for Transformers, which facilitates the learning of attention maps via a chain of residual convolutional neural networks. It obtains superior performance in various tasks in both CV and NLP domains.
Future works are considered in three aspects. First, we will apply evolving attention to more tasks and domains, such as object detection, question answering and time-series forecasting. Second, we aim to investigate other modules instead of convolutions to capture generic patterns in attention maps.
Last but not least, we would like to explore bi-directional decoders that benefit more from convolutions.

\bibliography{example_paper}
\bibliographystyle{icml2021}

\appendix
\section{Settings for reproduction}

\subsection{Image classification}
\begin{figure}[h]\setcounter{subfigure}{0}
	\centering
	\subfigure[Analysis of $\alpha$]{
        \centering
        \includegraphics[width=0.9\linewidth]{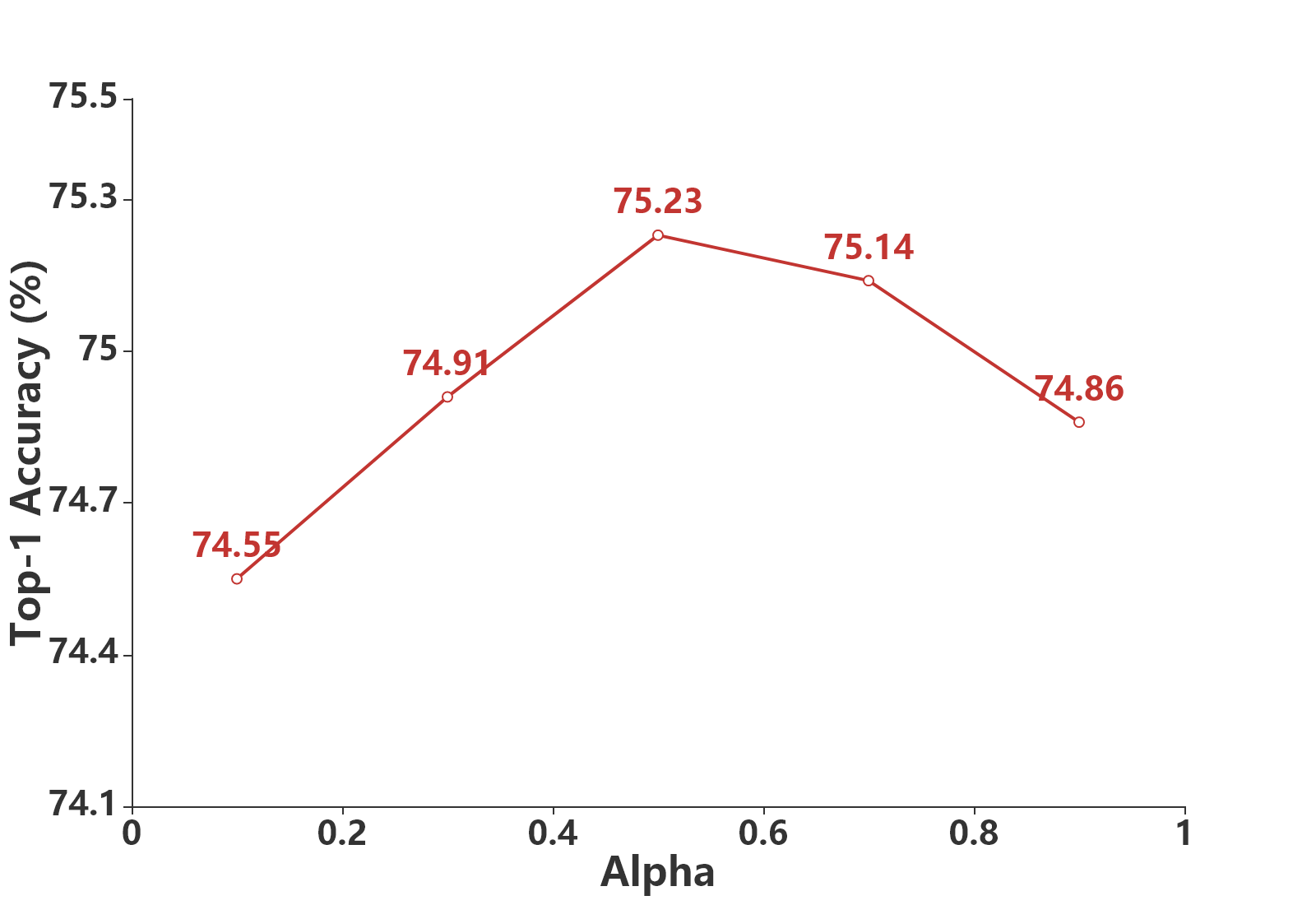}
        \label{fig:alpha}
    }
    \subfigure[Analysis of $\beta$]{
        \centering
        \includegraphics[width=0.9\linewidth]{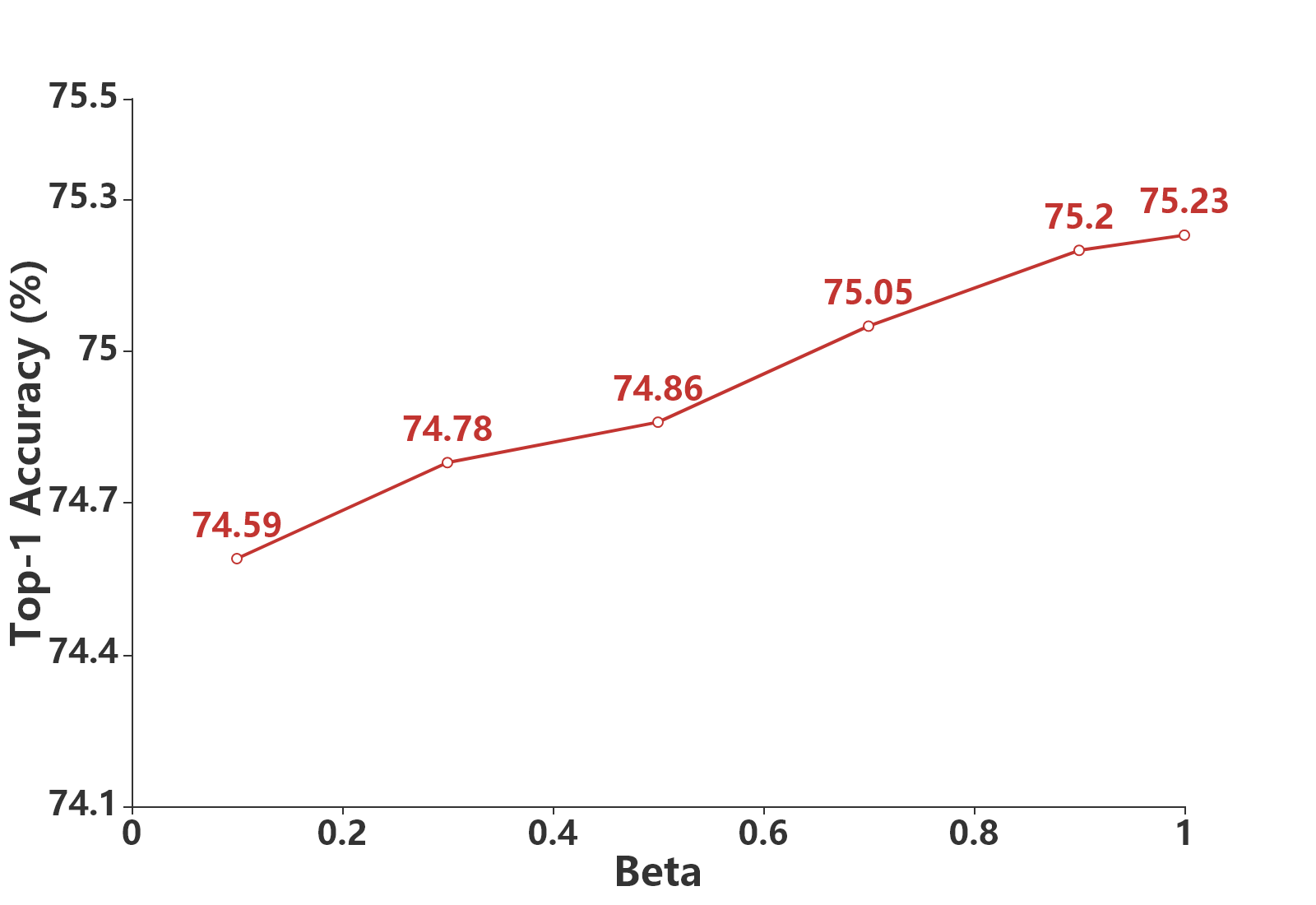}
        \label{fig:beta}
    }
    \caption{Hyper-parameter analysis for EA-AA-ResNet-34}
    \label{fig:hyper-parameter}
\end{figure}

We follow a common strategy described in \citet{szegedy2016rethinking} for data augmentation. For ResNet~\citep{he2016deep}, we adopt the implementation in tensorflow \footnote{\url{https://github.com/tensorflow/models/tree/v1.13.0}}. For AA-ResNet, we modify the ResNet by augmenting 3x3 convolution with self-attention. The implementation is from the official repository of AA-ResNet\footnote{\url{https://github.com/leaderj1001/Attention-Augmented-Conv2d}}, while we simply add the residual convolution module for EA-AA-ResNet. Concretely, we apply attention augmentation to each residual block in the last three stages -- when the shapes of activation maps become 28x28, 14x14 and 7x7. We adopt the same setting as AA-ResNet, e.g. $k=2$ and $v=0.2$. We refer to \cite{bello2019attention} for more details. We set $\alpha=0.5$ and $\beta=1.0$ by default, except for EA-AA-ResNet-152 and EA-AA-ResNet-101 where we set $\alpha = 0.7$. A hyper-parameter analysis for $\alpha$ and $\beta$ in EA-AA-ResNet-34 is shown in Figure \ref{fig:hyper-parameter}. We can see that the best result is achieved at $\alpha=0.5$ and $\beta=1.0$, which significantly outperforms the vanilla transformer (equivalent to $\alpha=0$ and $\beta=0$).

\begin{table*}[t]
\centering
\scalebox{0.85}{
    \renewcommand\arraystretch{1.1}
    \centering
    \begin{tabular}{llccccccc}
    \toprule
     \textbf{Model}&\textbf{Task} & \textbf{Training Epochs} & \textbf{Batch Size} & \textbf{Learning Rate} & \textbf{Adam Epsilon} & \textbf{Dropout Rate} & $\alpha$ & $\beta$\\
     \midrule
     \multirow{9}{*}{BERT-Base}
     & CoLA   & 3 & 8 & 2e-5 & 1e-8 & 0.1 & 0.2 & 0.1\\
     & SST-2  & 5 & 8 & 1e-5 & 1e-8 & 0.1 & 0.1 & 0.1\\
     & MRPC   & 2 & 8 & 2e-5 & 1e-8 & 0.1 & 0.1 & 0.2\\
     & STS-B  & 3 & 8 & 2e-5 & 1e-8 & 0.1 & 0.2 & 0.2\\
     & QQP    & 3 & 16 & 2e-5 & 1e-8 & 0.1 & 0.1 & 0.2\\
     & MNLI   & 3 & 16 & 2e-5 & 1e-8 & 0.1 & 0.1 & 0.2\\
     & QNLI   & 3 & 16 & 2e-5 & 1e-8 & 0.1 & 0.1 & 0.1\\
     & RTE    & 2 & 8 & 2e-5 & 1e-8 & 0.1 & 0.2 & 0.1\\
     & WNLI   & 2 & 8 & 2e-5 & 1e-8 & 0.1 & 0.2 & 0.2\\
     \multirow{9}{*}{BERT-Large}
     & CoLA   & 3 & 8 & 2e-5 & 1e-8 & 0.1 & 0.1 & 0.2\\
     & SST-2  & 5 & 8 & 1e-5 & 1e-8 & 0.1 & 0.1 & 0.2\\
     & MRPC   & 2 & 8 & 1e-5 & 1e-8 & 0.1 & 0.1 & 0.1\\
     & STS-B  & 3 & 8 & 2e-5 & 1e-8 & 0.1 & 0.1 & 0.2\\
     & QQP    & 3 & 16 & 1e-5 & 1e-8 & 0.1 & 0.1 & 0.1\\
     & MNLI   & 3 & 16 & 2e-5 & 1e-8 & 0.1 & 0.2 & 0.2\\
     & QNLI   & 3 & 16 & 2e-5 & 1e-8 & 0.1 & 0.2 & 0.2\\
     & RTE    & 3 & 16 & 1e-4 & 1e-8 & 0.1 & 0.1 & 0.2\\
     & WNLI   & 2 & 8 & 2e-5 & 1e-8 & 0.1 & 0.2 & 0.2\\
     \multirow{9}{*}{RoBERTa-Large}
     & CoLA   & 3 & 8 & 2e-5 & 1e-8 & 0.1 & 0.2 & 0.1\\
     & SST-2  & 5 & 8 & 1e-5 & 1e-8 & 0.1 & 0.1 & 0.2\\
     & MRPC   & 3 & 16 & 1e-5 & 1e-8 & 0.1 & 0.1 & 0.2\\
     & STS-B  & 3 & 16 & 2e-5 & 1e-8 & 0.1 & 0.4 & 0.2\\
     & QQP    & 3 & 16 & 2e-5 & 1e-8 & 0.1 & 0.1 & 0.1\\
     & MNLI   & 5 & 16 & 1e-5 & 1e-8 & 0.1 & 0.4 & 0.1\\
     & QNLI   & 5 & 16 & 1e-5 & 1e-8 & 0.1 & 0.4 & 0.2\\
     & RTE    & 2 & 8 & 1e-4 & 1e-8 & 0.1 & 0.2 & 0.4\\
     & WNLI   & 2 & 8 & 2e-5 & 1e-8 & 0.1 & 0.2 & 0.1\\
     \multirow{9}{*}{T5-Base}
     & CoLA   & 3 & 8 & 2e-5 & 1e-8 & 0.1 & 0.2 & 0.2\\
     & SST-2  & 3 & 8 & 2e-5 & 1e-8 & 0.1 & 0.1 & 0.2\\
     & MRPC   & 3 & 8 & 1e-4 & 1e-8 & 0.1 & 0.1 & 0.1\\
     & STS-B  & 3 & 16 & 2e-5 & 1e-8 & 0.1 & 0.1 & 0.1\\
     & QQP    & 3 & 16 & 2e-5 & 1e-8 & 0.1 & 0.1 & 0.2\\
     & MNLI   & 5 & 16 & 1e-5 & 1e-8 & 0.1 & 0.1 & 0.1\\
     & QNLI   & 5 & 16 & 1e-5 & 1e-8 & 0.1 & 0.2 & 0.1\\
     & RTE    & 3 & 16 & 1e-5 & 1e-8 & 0.1 & 0.4 & 0.2\\
     & WNLI   & 2 & 8 & 2e-5 & 1e-8 & 0.1 & 0.2 & 0.2\\
     \bottomrule
    \end{tabular}
    }
    \caption{Detailed hyper-parameter settings for GLUE benchmark.}
    \label{tab:gluepara}
\end{table*}

\subsection{Natural language understanding}

As introduced in \cite{devlin2018bert}, BERT-Base and EA-BERT-Base have 12 layers and 12 attention heads with hidden dimension 768. BERT-large and EA-BERT-large have 24 layers and 16 heads, while hidden dimension for each intermediate layer is set as 1024. The hidden dimension of the final fully-connected layer before softmax is set to be 2000. 
We download the official checkpoints of BERT-Base\footnote{\url{https://storage.googleapis.com/bert_models/2018_10_18/uncased_L-12_H-768_A-12.zip}} and BERT-Large\footnote{\url{https://storage.googleapis.com/bert_models/2018_10_18/uncased_L-24_H-1024_A-16.zip}}, and initialize the additional parameters for EA-BERT-Base and EA-BERT-Large randomly.

We also conduct a set of experiments with RoBERTa-Large~\citep{liu2019roberta} and T5-Base~\citep{raffel2019exploring}. We apply the idea of evolving attention to the network of these models. RoBERTa-Large has 24 layers with 16 attention heads. The total hidden size of all heads is 1024, and the hidden dimension of the final fully-connected layer is 4096. We use NLP library implemented by the huggingface team~\cite{Wolf2019HuggingFacesTS} to implement the base version of T5, which has 220 million parameter. We download the official pre-trained checkpoints for RoBERTa-Large\footnote{\url{https://dl.fbaipublicfiles.com/fairseq/models/roberta.large.tar.gz}} and T5-Base\footnote{\url{https://console.cloud.google.com/storage/browser/t5-data/pretrained_models/base/}} for fine-tuning.

We adopt the Adam optimizer~\citep{kingma2014adam} with epsilon 1e-8.
The dropout rate is set as 0.1 empirically. We use grid search to optimize the values of hyper-parameters on the validation set. We search the learning rate in \{1e-4, 1e-5, 2e-5\}, batch size in \{8, 16\}, training epochs in \{2, 3, 5\}, $\alpha$ in \{0.1, 0.2, 0.4\} and $\beta$ in \{0.1, 0.2, 0.4\}. 
The specific hyper-parameters for each task are listed in Table \ref{tab:gluepara}.

\begin{table*}[t]
\centering
\scalebox{0.8}{
    \renewcommand\arraystretch{1.1}
    \centering
    \begin{tabular}{llcccccc}
    \toprule
     \textbf{Model} & \textbf{Task} & \textbf{Number of GPU} & \textbf{Accumulative Steps} & \textbf{Learning Rate} & \textbf{Dropout Rate} & $\alpha$ & $\beta$\\
     \midrule
     \multirow{3}{*}{EA-Transformer-Lite} & De-En & 1 & 1 & 1e-3 & 0.2 & 0.1 & 0.1 \\
     & En-De & 8 & 16 & 1e-3 & 0.3 & 0.1 & 0.1 \\
     & En-Fr & 8 & 16 & 1e-3 & 0.1 & 0.1 & 0.1 \\
     \multirow{3}{*}{EA-Transformer-Base} & De-En & 1 & 1 & 1e-3 & 0.2 & 0.5 & 0.1 \\
       & En-De & 8 & 16 & 1e-3 & 0.3 & 0.5 & 0.1 \\
      & En-Fr & 8 & 16 & 1e-3 & 0.1 & 0.5 & 0.1 \\
     \bottomrule
    \end{tabular}
    }
    \caption{Detailed hyper-parameter settings for machine translation.}
    \label{tab:mt_hyper}
\end{table*}

\subsection{Machine Translation}
We train the machine translation tasks using Adam optimizer~\citep{kingma2014adam} with $\beta_1 = 0.9$, $\beta_2 = 0.98$ and an inverse square root learning rate scheduling with linear warmup. The learning rate is 1e-3 and the warmup step is set to be 4000. Also, we use label smoothing $\epsilon = 0.1$. we apply early stopping to the training procedure with a patience of 5 epochs. In the evaluation phase, we average the final 10 checkpoints and conduct beam search with size 5. We adopt absolute positional encoding according to the original implementation~\citep{vaswani2017attention}. Other hyper-parameters are optimized by grid search on the validation set and reported in Table \ref{tab:mt_hyper}.

\begin{table}[t]
\centering
\scalebox{0.76}{
    \renewcommand\arraystretch{1.1}
    \begin{tabular}{lcccc}
    \toprule
     & \textbf{Perf.$\uparrow$} & \textbf{AUPRC$\uparrow$} & \textbf{Comp.$\uparrow$} & \textbf{Suff.$\downarrow$} \\
     \midrule
     \textbf{Movie Reviews}\\
     BERT+LSTM - Attention & 0.970 & 0.417 & 0.129 & 0.097\\
     BERT+LSTM - EA-Attention & 0.970 & \textbf{0.435} & 0.142 & \textbf{0.084}\\
     BERT+LSTM - Lime & 0.970 & 0.280 & 0.187 & 0.093\\
     EA-BERT+LSTM -Lime & \textbf{0.975} & 0.313 & \textbf{0.194} & 0.089\\
     \midrule
     \textbf{FEVER}\\
     BERT+LSTM - Attention & 0.870 & 0.235 & 0.037 & 0.122\\
     BERT+LSTM - EA-attention & 0.870 & 0.238 & 0.078 & 0.097\\
     BERT+LSTM - Lime & 0.870 & 0.291 & 0.212 & \textbf{0.014}\\
     EA-BERT+LSTM -Lime & \textbf{0.886} & \textbf{0.307} & \textbf{0.236} & \textbf{0.014}\\
     \midrule
    \textbf{MultiRC}\\
     BERT+LSTM - Attention & 0.655 & 0.244 & 0.036 & 0.052\\
     BERT+LSTM - EA-Attention & 0.655 & \textbf{0.251} & 0.054 & 0.041\\
     BERT+LSTM - Lime & 0.655 & 0.208 & 0.213 & -0.079\\
     EA-BERT+LSTM -Lime & \textbf{0.674} & 0.221 & \textbf{0.241} & \textbf{-0.089}\\
     \midrule
     \textbf{CoS-E}\\
     BERT+LSTM - Attention & 0.487 & 0.606 & 0.080 & 0.217\\
     BERT+LSTM - EA-Attention & 0.487 & \textbf{0.610} & 0.113 & 0.189 \\
     BERT+LSTM - Lime & 0.487 & 0.544 & 0.223 & 0.143\\
     EA-BERT+LSTM -Lime & \textbf{0.491} & 0.552 & \textbf{0.231} & \textbf{0.140}\\
     \midrule
     \textbf{e-SNLI}\\
     BERT+LSTM - Attention & 0.960 & 0.395 & 0.105 & 0.583\\
     BERT+LSTM - EA-Attention & 0.960 & 0.399 & 0.177 & 0.396\\
     BERT+LSTM - Lime & 0.960 & 0.513 & 0.437 & 0.389\\
     EA-BERT+LSTM -Lime & \textbf{0.969} & \textbf{0.534} & \textbf{0.445} & \textbf{0.368}\\
    \bottomrule
    \end{tabular}
}
 \caption{Comparison of different text representation models and rationale generation methods on ERASER benchmark. ``Perf." is accuracy (CoS-E) or F1 (others), AUPRC means Area Under the Precision Recall Curve; ``Comp." and ``Suff." denote comprehensiveness and sufficiency metrics respectively.}
 \label{app:rationale}
\end{table}

\section{Analysis}
\subsection{Quality of Image Attention}
We select the 16th, 17th and 18th attention layers in the AA-ResNet-34 and EA-AA-ResNet34 networks for analysis. The attention maps from these layers have a shape of $14 \times 14 \times 8$, where 14 is the image length after pooling and 8 is the number of heads. Then, we send the attention maps directly as inputs to another CNN model for classification, and the original labels are used for training and evaluation. The goal is to quantify the effectiveness of attention maps learned by different models. If an attention map retains major structures of the original object, the classification accuracy should be higher. We adopt a 12-layer DenseNet~\cite{huang2017densely} for attention map classification, while the shape is pooled to $7 \times 7$ and $4 \times 4$ after the 4th and 7th layers respectively. The hidden dimension is set as 256 initially and doubled after each pooling operation. The models are trained by 30 epochs with cosine learning rate decay started by 0.05 and ended by 0.0001.

\subsection{Interpretability}
In the context of modern neural models, attention mechanisms learn to assign soft weights to token representations, thus one can extract highly weighted tokens as rationales. In other words, if the self-attention mechanism generates better attention scores, it will achieve better performance on ERASER, a benchmark designed to evaluate the interpretability of natural language understanding models.
We list the experimental results on the ERASER benchmark in Table \ref{app:rationale}. It should be noted that higher comprehensiveness scores and lower sufficiency scores are desired. First, we consider a text representation model that passes tokens through BERT and a bidirectional LSTM. Based on the same text representation, we use different methods for rationale generation. Here, the models are the same, so the downstream performance is equivalent. According to the experimental results, the rationales generated using evolving attention are more accurate than the ones generated by vanilla attention.
Next, we replace vanilla BERT with EA-BERT as the text representation model and utilize a state-of-the-art rationale generation method, Lime~\citep{ribeiro2016should}. Experimental results show that EA-BERT improves the performances of downstream tasks and generates better rationales simultaneously.

\section{Case Study}
In order to get insight into the evolving attention mechanism, we visualize exemplar attention maps for both text and image inputs and find some interesting evidences.

\subsection{Image attention}

\begin{figure*}[h]
    \begin{center}
        \includegraphics[width=1.0\linewidth]{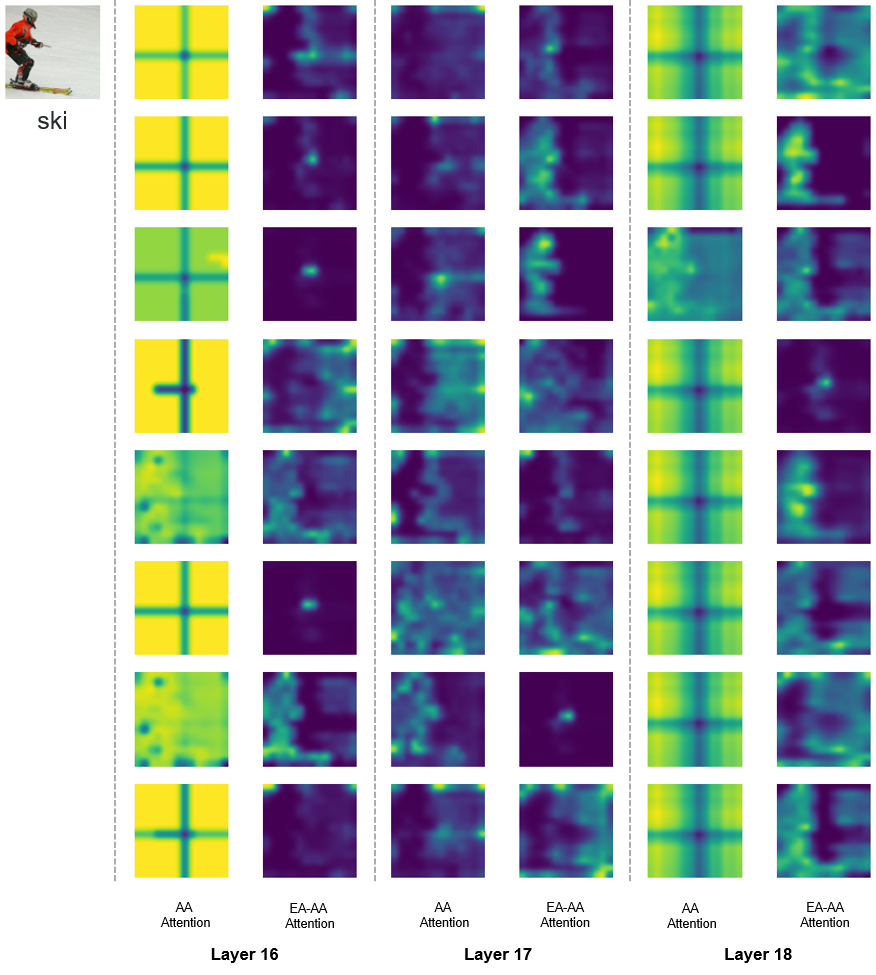}
    \end{center}
      \caption{Attention map visualization for an image classification example}
      \label{fig:image_example_1}
\end{figure*}

\begin{figure*}[h]
    \begin{center}
        \includegraphics[width=1.0\linewidth]{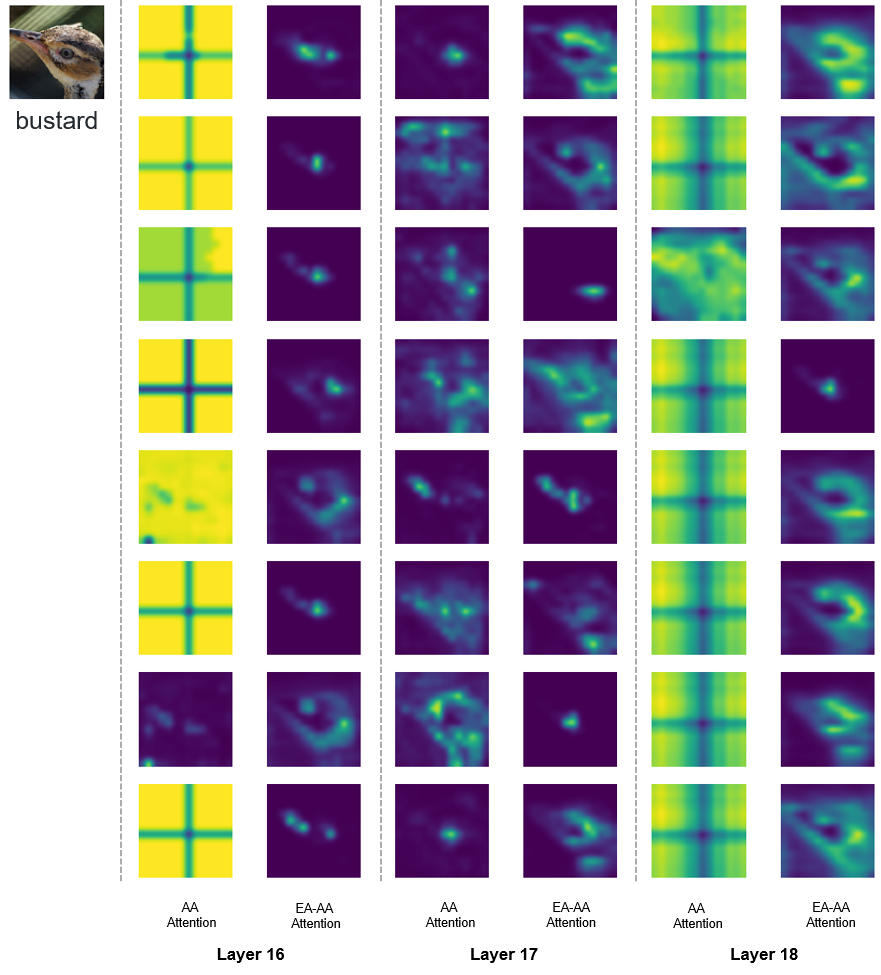}
    \end{center}
     \caption{Attention map visualization for an image classification example}
     \label{fig:image_example_2}
\end{figure*}

\begin{figure*}[h]
    \begin{center}
        \includegraphics[width=1.0\linewidth]{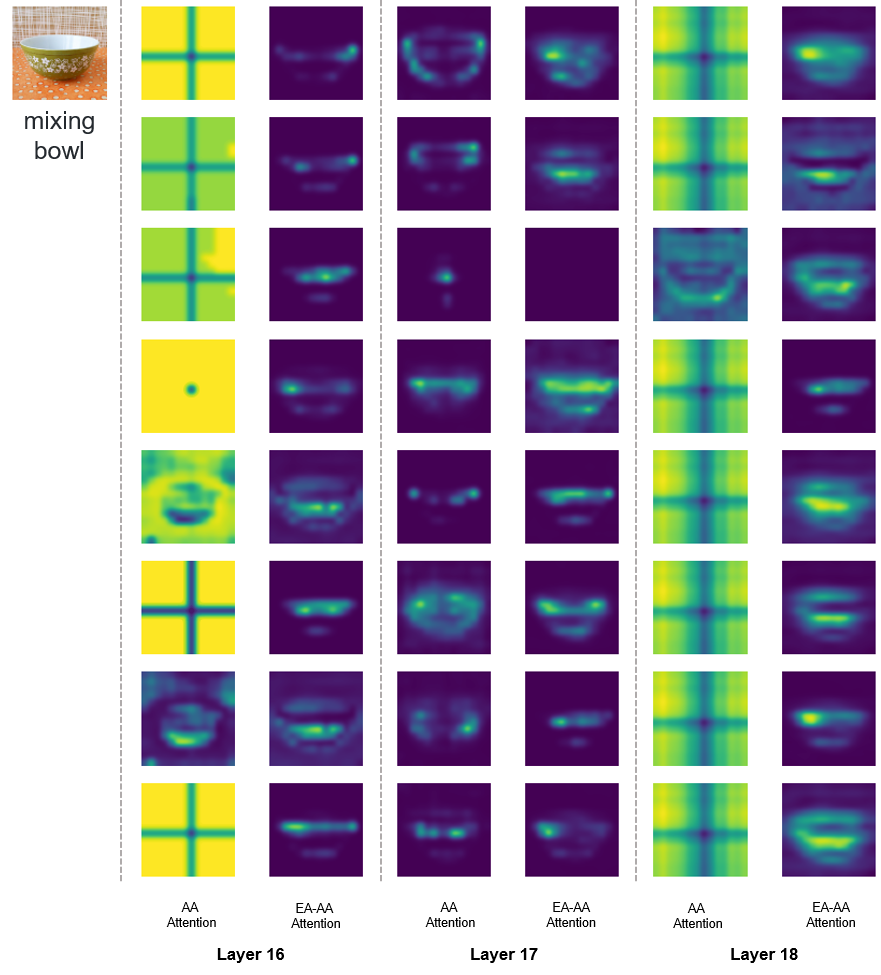}
    \end{center}
     \caption{Attention map visualization for an image classification example}
     \label{fig:image_example_3}
\end{figure*}

\begin{figure*}[h]
    \begin{center}
        \includegraphics[width=1.0\linewidth]{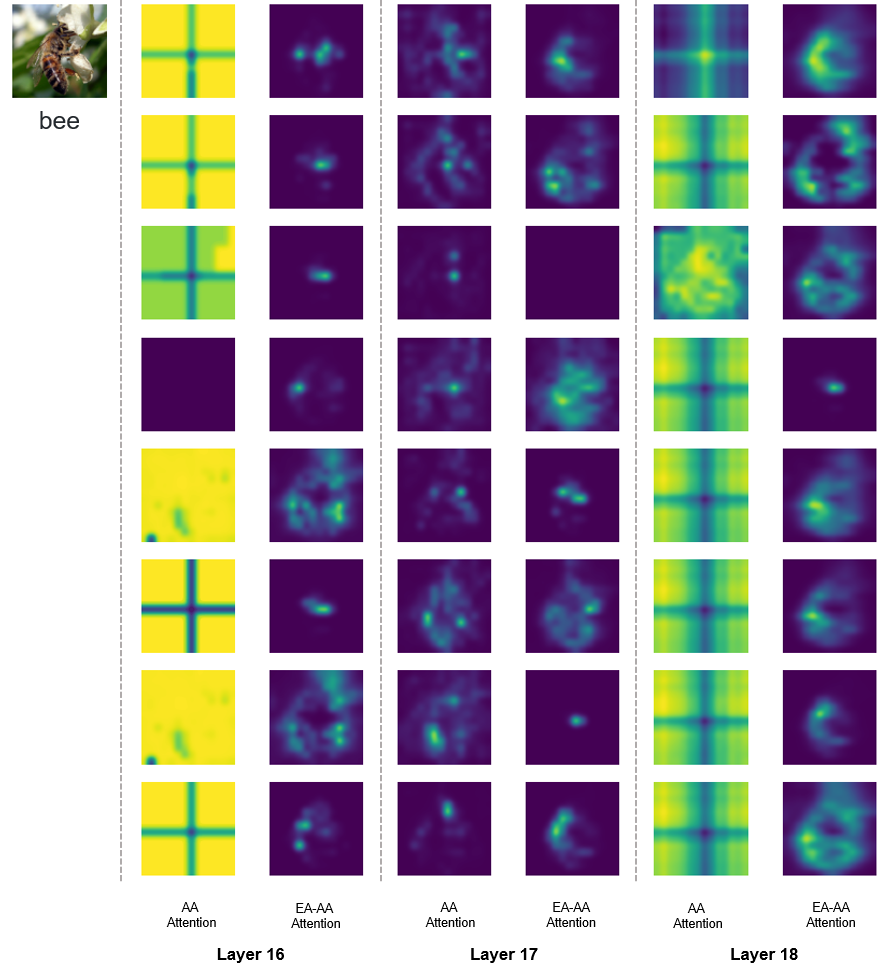}
    \end{center}
     \caption{Attention map visualization for an image classification example}
     \label{fig:image_example_4}
\end{figure*}

In Figure \ref{fig:image_example_1}-\ref{fig:image_example_4}, we compare the attention maps of AA-ResNet-34 and EA-AA-ResNet-34 for ImageNet classification. Compared to AA-ResNet, our proposed convolution-based evolving attention mechanism captures better global information and at the same time emphasizes on the important local information. Specifically, the residual connections and convolutional inductive bias assist the self-attention mechanism to depict a more clear outline. As shown by the visualized examples, AA-ResNet fails to compute a explainable attention map for some layers. In contrast, with the help of residual convolutions, EA-AA-ResNet successfully identifies the objects in images in an evolving process.

\subsection{Text Attention}

We choose BERT-Base and EA-BERT-Base models for comparison on the CoLA dataset, a task of judging the grammatical correctness of a sentence. We select the sentence ``\textit{Mary tried John to go abroad.}" for a case study. Obviously, this sentence is grammatically wrong, and a model should capture the error part ``\textit{tried John to}" in order to give the correct answer.

\begin{figure*}[t]\setcounter{subfigure}{0}
	\centering
	\subfigure[BERT \#2]{
        \begin{minipage}[t]{0.25\linewidth}
        \centering
        \includegraphics[width=\linewidth]{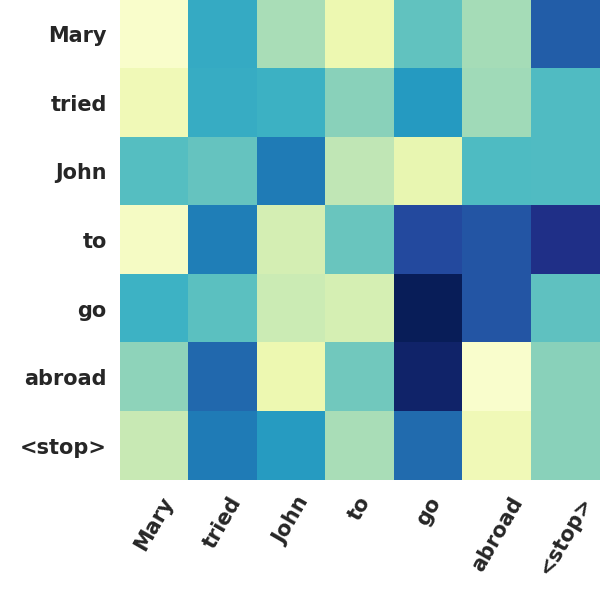}
        \label{fig:vis_raw2}
    \end{minipage}%
    }\subfigure[EA-BERT \#2]{
        \begin{minipage}[t]{0.25\linewidth}
        \centering
        \includegraphics[width=\linewidth]{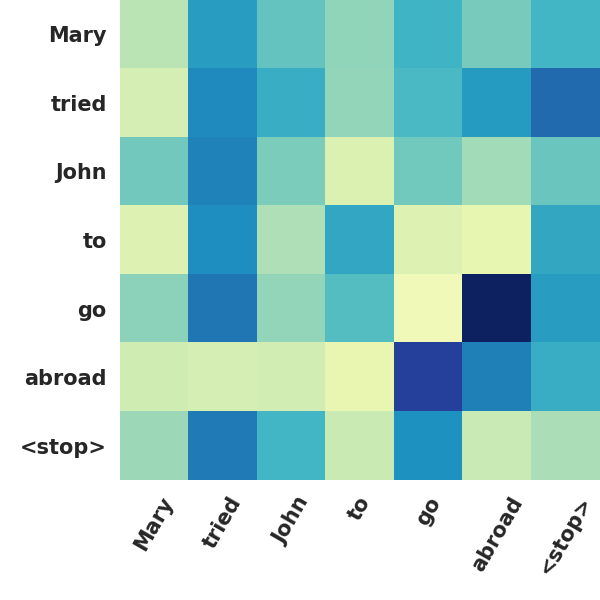}
        \label{fig:vis_conv2}
    \end{minipage}%
    }\subfigure[Convolution \#2]{
        \begin{minipage}[t]{0.25\linewidth}
        \centering
        \includegraphics[width=\linewidth]{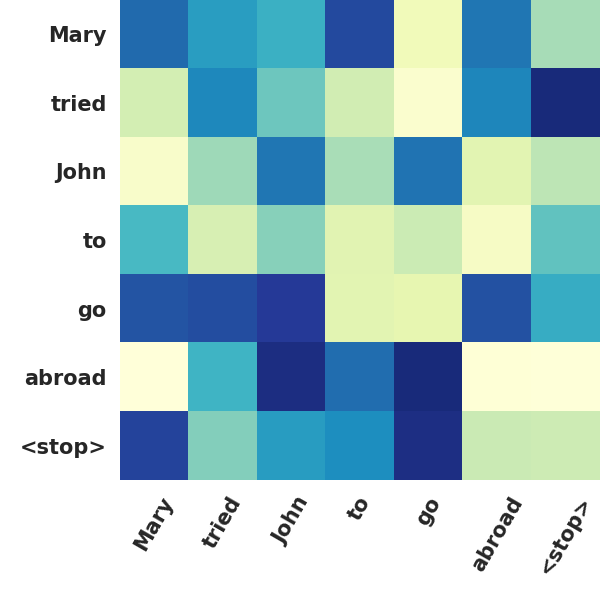}
        \label{fig:vis_conv2_alpha}
    \end{minipage}%
    }\subfigure[Self-Attention \#2]{
        \begin{minipage}[t]{0.25\linewidth}
        \centering
        \includegraphics[width=\linewidth]{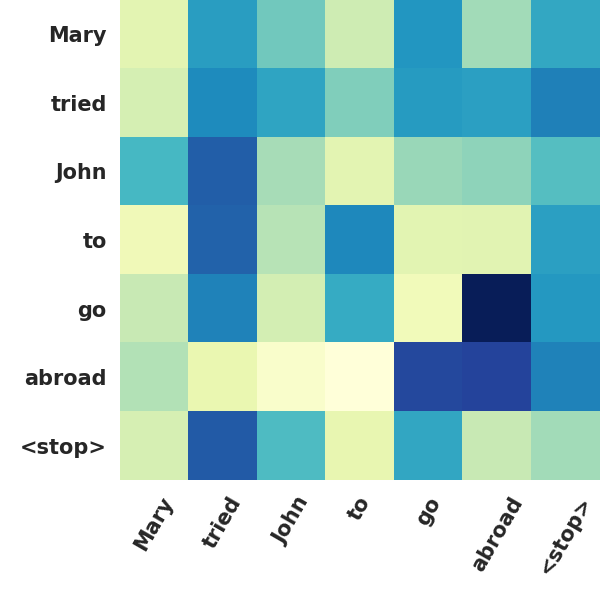}
        \label{fig:vis_conv2_1-alpha}
    \end{minipage}%
    }
    \label{case_study_layer2}

	\subfigure[BERT \#11]{
        \begin{minipage}[t]{0.25\linewidth}
        \centering
        \includegraphics[width=\linewidth]{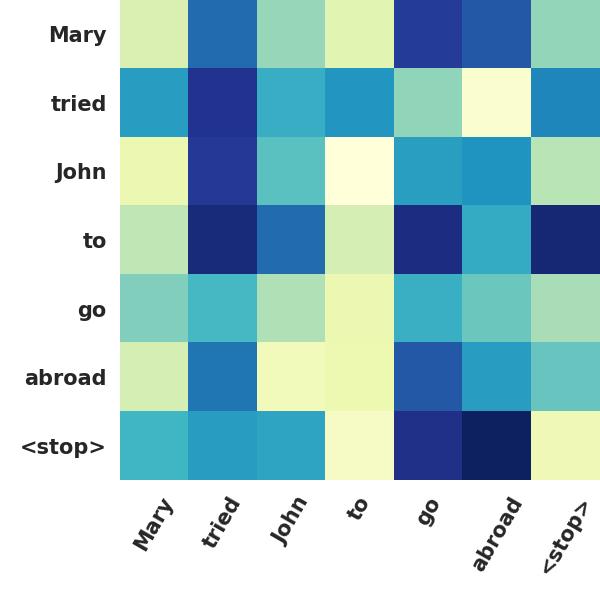}
        \label{fig:vis_raw11}
    \end{minipage}%
    }\subfigure[EA-BERT \#11]{
        \begin{minipage}[t]{0.25\linewidth}
        \centering
        \includegraphics[width=\linewidth]{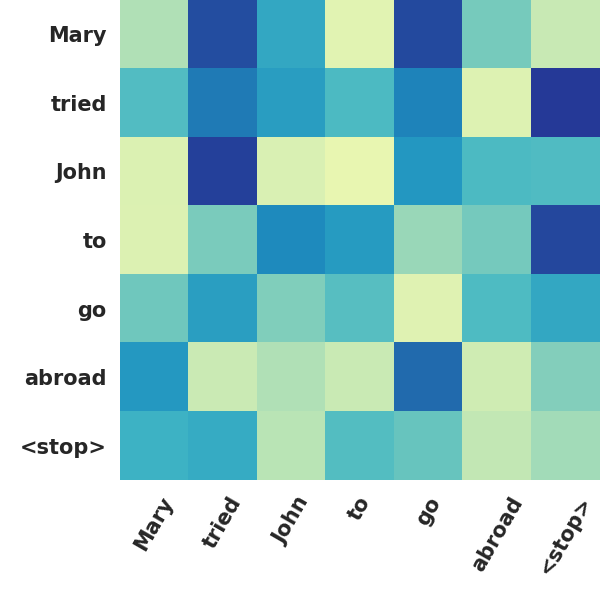}
        \label{fig:vis_conv11}
    \end{minipage}%
    }\subfigure[Convolution \#11]{
        \begin{minipage}[t]{0.25\linewidth}
        \centering
        \includegraphics[width=\linewidth]{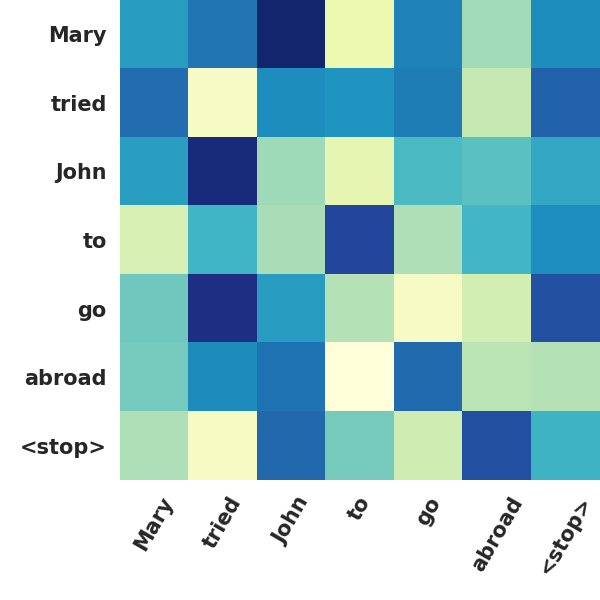}
        \label{fig:vis_conv11_alpha}
    \end{minipage}%
    }\subfigure[Self-Attention \#11]{
        \begin{minipage}[t]{0.25\linewidth}
        \centering
        \includegraphics[width=\linewidth]{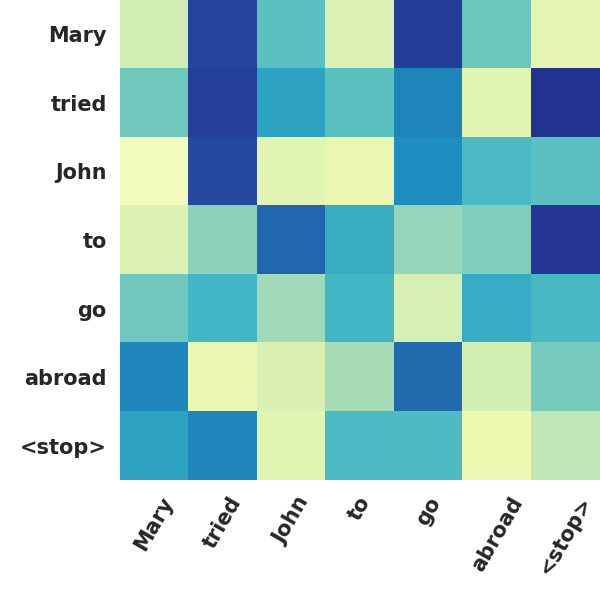}
        \label{fig:vis_conv11_1-alpha}
    \end{minipage}%
    }
    \label{case_study_layer11}

	\subfigure[BERT \#12]{
        \begin{minipage}[t]{0.25\linewidth}
        \centering
        \includegraphics[width=\linewidth]{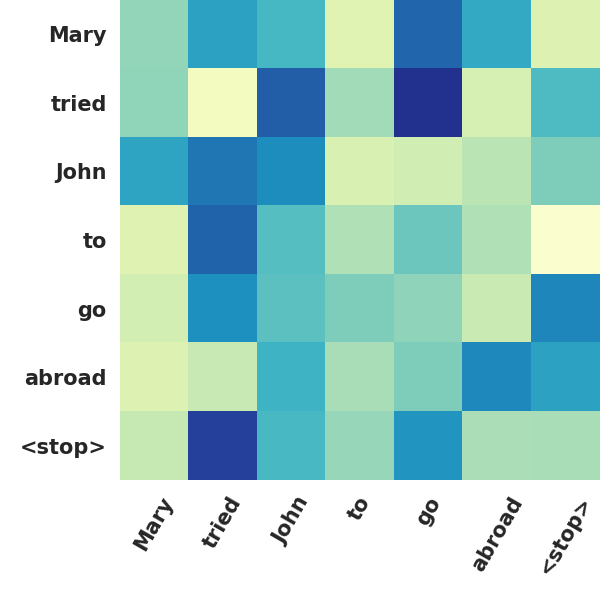}
        \label{fig:vis_raw12}
    \end{minipage}%
    }\subfigure[EA-BERT \#12]{
        \begin{minipage}[t]{0.25\linewidth}
        \centering
        \includegraphics[width=\linewidth]{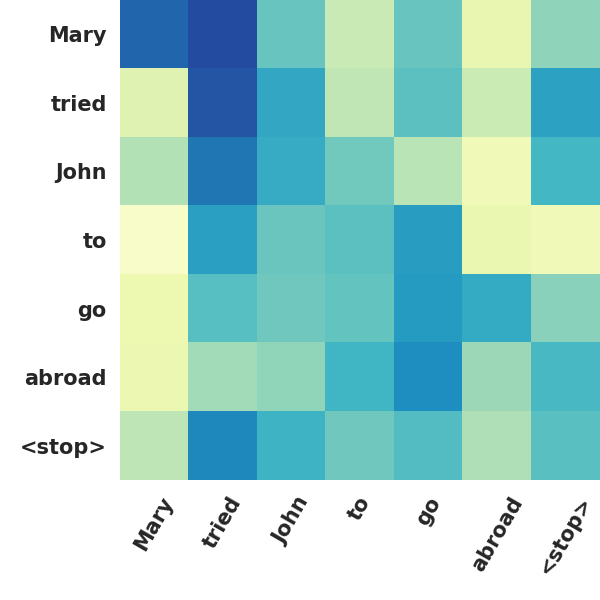}
        \label{fig:vis_conv12}
    \end{minipage}%
    }\subfigure[Convolution \#12]{
        \begin{minipage}[t]{0.25\linewidth}
        \centering
        \includegraphics[width=\linewidth]{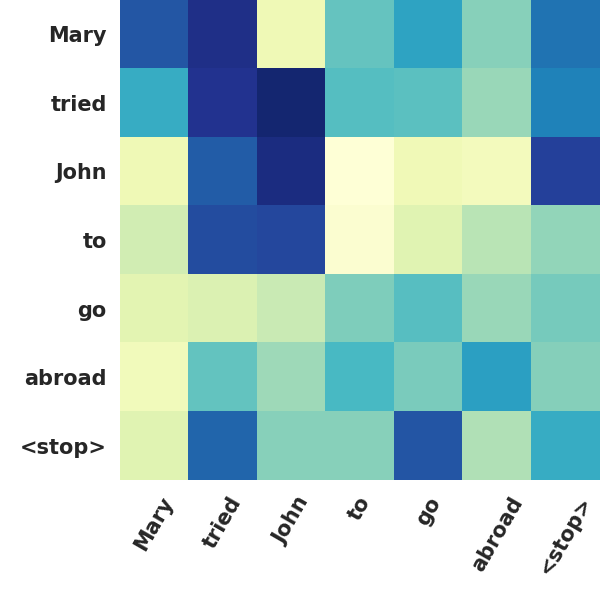}
        \label{fig:vis_conv12_alpha}
    \end{minipage}%
    }\subfigure[Self-Attention \#12]{
        \begin{minipage}[t]{0.25\linewidth}
        \centering
        \includegraphics[width=\linewidth]{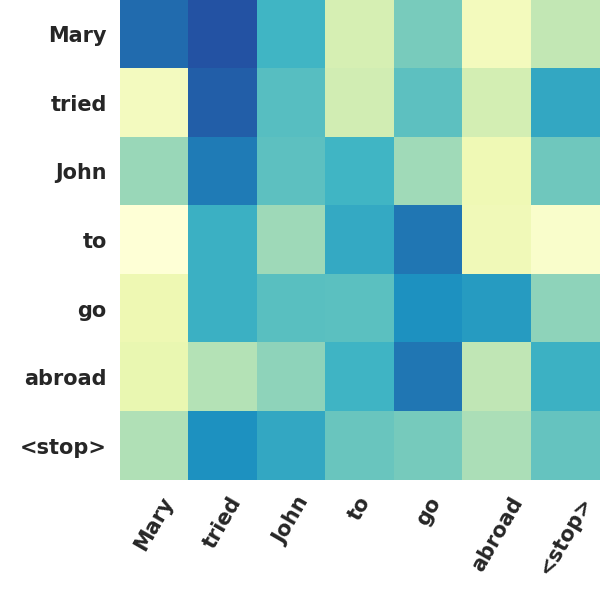}
        \label{fig:vis_conv12_1-alpha}
    \end{minipage}%
    }
    \caption{Attention maps of layer \#2, \#11 and \#12 for ``\textit{Mary tried John to go abroad.}"}
    \label{case_study_layer12}
    \label{fig:case_study}
\end{figure*}

In Figure \ref{fig:case_study}, we visualize related attention maps for three layers (\#2, \#11 and \#12) in BERT-Base and EA-BERT-Base models. The second layer is the first layer that utilizes residual attention, and \#11 and \#12 are the last two layers. For each layer, we first show the attention maps from vanilla BERT and EA-BERT in the first and second columns respectively, then the convolution-based attention and self-attention maps are visualized in the third and fourth column. It should be noted that the second column is the linear fusion result of the third column and the fourth column.

Consider layer \#2, both BERT (Figure \ref{fig:vis_raw2}) and EA-BERT (Figure \ref{fig:vis_conv2}) pay major attentions on the verb phrase ``\textit{go abroad}". As shown in Figure \ref{fig:vis_conv2}, EA-BERT puts additional stress on the relation between word ``\textit{tried}" and the stop sign. This is reasonable because the stop sign is responsible of capturing sentence-level semantics and ``\textit{tried}" is a key word leading to the grammatical error. As shown in Figure \ref{fig:vis_conv2_alpha}, the attention on this part actually comes from the convolution-based module, which is sometimes complementary to the self-attention map.

In order to ensure that the information obtained by the convolution is beneficial, we visualize the last attention layer (\#12) which is the closest to the output (see Figure \ref{case_study_layer12}(i-l)). In Figure \ref{fig:vis_raw12}, we can observe that BERT-Base still focuses on verbs and stop signs in the very last layer of transformer. The attention to the wrong phrase ``\textit{tried John}" is still relatively weak, leading to a mis-classification result. In contrast, the attention scores between ``\textit{tried}" and ``\textit{John}" become obvious in EA-BERT (Figure \ref{fig:vis_conv12}), largely owning to the convolutional attention map illustrated in Figure \ref{fig:vis_conv12_alpha}.

We also visualize the attention maps of the \#11 layer, which serves as input to the \#12 layer.
To analysis the evolution of attention maps, we compare the differences between Figure \ref{fig:vis_conv11} and Figure \ref{fig:vis_conv12_alpha}, as the latter is the evolved attention map taking the former as input. We find that the convolutional module helps to reason about the important word relations based on the previous attention maps. Specifically, it weakens the attention scores of the correct parts and raises higher scores for the wrong parts.
As illustrated in Figure \ref{fig:vis_conv12_alpha}, the attention scores are significant in the upper left corner of the matrix where the error occurs. In this way, the error is fully captured in the final representation layer, helping EA-BERT to generate a correct answer.

\end{document}